\documentclass{article}
\usepackage{arxiv}

\usepackage[english]{babel}
\usepackage[utf8x]{inputenc}
\usepackage[T1]{fontenc}
\usepackage{bm}
\usepackage{amssymb}
\usepackage{epsfig}
\usepackage{url}
\usepackage{amsmath}
\usepackage{amsthm}
\usepackage{booktabs}
\usepackage{multicol}
\usepackage{amssymb}
\usepackage{amsmath}
\usepackage{graphicx}
\usepackage[colorinlistoftodos]{todonotes}

\newcommand{\ry}{\mathrm{y}}

\newcommand{\bbE}{\mathbb{E}}

\newcommand{\vx}{\mathbf{x}}

\newcommand{\rx}{\mathrm{x}}

\newcommand{\elle}{\mathit{l}}
\newcommand{\base}{\mathit{g}}

\newcommand{\lrsb}[1]{\left[#1\right]}

\newcommand{\bfXr}{\mathbf{X}}                        

\newcommand{\underover}[3]{\overset{#2}{\underset{#3}{#1}}}

\newcommand{\nmathbf}{\bm}

\def\bfx{\nmathbf x}

\def\bfbeta   {\nmathbf \beta}

\def\bfmu     {\nmathbf \mu}

\def\boldfacefake#1{\kern-4pt
    \hbox{ \mathsurround=0pt
    \hbox to 0.4pt{$#1$\hss}\hbox to 0.4pt{$#1$\hss}\hbox {$#1$}}}

\newcommand{\be}{\begin{eqnarray}}
\newcommand{\ee}{\end{eqnarray}}
\newcommand{\ba}{\begin{eqnarray*}}
\newcommand{\ea}{\end{eqnarray*}}

\newcommand{\reals}{\mbox{\rm I\kern-.20em R}}
\newcommand{\sreals}{\mbox{\small \rm I\kern-.20em R}}

\definecolor{AliceBlue}{rgb}{0.94,0.97,1.00}
\definecolor{AntiqueWhite1}{rgb}{1.00,0.94,0.86}
\definecolor{AntiqueWhite2}{rgb}{0.93,0.87,0.80}
\definecolor{AntiqueWhite3}{rgb}{0.80,0.75,0.69}
\definecolor{AntiqueWhite4}{rgb}{0.55,0.51,0.47}
\definecolor{AntiqueWhite}{rgb}{0.98,0.92,0.84}
\definecolor{BlanchedAlmond}{rgb}{1.00,0.92,0.80}
\definecolor{BlueViolet}{rgb}{0.54,0.17,0.89}
\definecolor{CadetBlue1}{rgb}{0.60,0.96,1.00}
\definecolor{CadetBlue2}{rgb}{0.56,0.90,0.93}
\definecolor{CadetBlue3}{rgb}{0.48,0.77,0.80}
\definecolor{CadetBlue4}{rgb}{0.33,0.53,0.55}
\definecolor{CadetBlue}{rgb}{0.37,0.62,0.63}
\definecolor{CornflowerBlue}{rgb}{0.39,0.58,0.93}
\definecolor{DarkBlue}{rgb}{0.00,0.00,0.55}
\definecolor{DarkCyan}{rgb}{0.00,0.55,0.55}
\definecolor{DarkGoldenrod1}{rgb}{1.00,0.73,0.06}
\definecolor{DarkGoldenrod2}{rgb}{0.93,0.68,0.05}
\definecolor{DarkGoldenrod3}{rgb}{0.80,0.58,0.05}
\definecolor{DarkGoldenrod4}{rgb}{0.55,0.40,0.03}
\definecolor{DarkGoldenrod}{rgb}{0.72,0.53,0.04}
\definecolor{DarkGray}{rgb}{0.66,0.66,0.66}
\definecolor{DarkGreen}{rgb}{0.00,0.39,0.00}
\definecolor{DarkGrey}{rgb}{0.66,0.66,0.66}
\definecolor{DarkKhaki}{rgb}{0.74,0.72,0.42}
\definecolor{DarkMagenta}{rgb}{0.55,0.00,0.55}
\definecolor{DarkOliveGreen1}{rgb}{0.79,1.00,0.44}
\definecolor{DarkOliveGreen2}{rgb}{0.74,0.93,0.41}
\definecolor{DarkOliveGreen3}{rgb}{0.64,0.80,0.35}
\definecolor{DarkOliveGreen4}{rgb}{0.43,0.55,0.24}
\definecolor{DarkOliveGreen}{rgb}{0.33,0.42,0.18}
\definecolor{DarkOrange1}{rgb}{1.00,0.50,0.00}
\definecolor{DarkOrange2}{rgb}{0.93,0.46,0.00}
\definecolor{DarkOrange3}{rgb}{0.80,0.40,0.00}
\definecolor{DarkOrange4}{rgb}{0.55,0.27,0.00}
\definecolor{DarkOrange}{rgb}{1.00,0.55,0.00}
\definecolor{DarkOrchid1}{rgb}{0.75,0.24,1.00}
\definecolor{DarkOrchid2}{rgb}{0.70,0.23,0.93}
\definecolor{DarkOrchid3}{rgb}{0.60,0.20,0.80}
\definecolor{DarkOrchid4}{rgb}{0.41,0.13,0.55}
\definecolor{DarkOrchid}{rgb}{0.60,0.20,0.80}
\definecolor{DarkRed}{rgb}{0.55,0.00,0.00}
\definecolor{DarkSalmon}{rgb}{0.91,0.59,0.48}
\definecolor{DarkSeaGreen1}{rgb}{0.76,1.00,0.76}
\definecolor{DarkSeaGreen2}{rgb}{0.71,0.93,0.71}
\definecolor{DarkSeaGreen3}{rgb}{0.61,0.80,0.61}
\definecolor{DarkSeaGreen4}{rgb}{0.41,0.55,0.41}
\definecolor{DarkSeaGreen}{rgb}{0.56,0.74,0.56}
\definecolor{DarkSlateBlue}{rgb}{0.28,0.24,0.55}
\definecolor{DarkSlateGray1}{rgb}{0.59,1.00,1.00}
\definecolor{DarkSlateGray2}{rgb}{0.55,0.93,0.93}
\definecolor{DarkSlateGray3}{rgb}{0.47,0.80,0.80}
\definecolor{DarkSlateGray4}{rgb}{0.32,0.55,0.55}
\definecolor{DarkSlateGray}{rgb}{0.18,0.31,0.31}
\definecolor{DarkSlateGrey}{rgb}{0.18,0.31,0.31}
\definecolor{DarkTurquoise}{rgb}{0.00,0.81,0.82}
\definecolor{DarkViolet}{rgb}{0.58,0.00,0.83}
\definecolor{DeepPink1}{rgb}{1.00,0.08,0.58}
\definecolor{DeepPink2}{rgb}{0.93,0.07,0.54}
\definecolor{DeepPink3}{rgb}{0.80,0.06,0.46}
\definecolor{DeepPink4}{rgb}{0.55,0.04,0.31}
\definecolor{DeepPink}{rgb}{1.00,0.08,0.58}
\definecolor{DeepSkyBlue1}{rgb}{0.00,0.75,1.00}
\definecolor{DeepSkyBlue2}{rgb}{0.00,0.70,0.93}
\definecolor{DeepSkyBlue3}{rgb}{0.00,0.60,0.80}
\definecolor{DeepSkyBlue4}{rgb}{0.00,0.41,0.55}
\definecolor{DeepSkyBlue}{rgb}{0.00,0.75,1.00}
\definecolor{DimGray}{rgb}{0.41,0.41,0.41}
\definecolor{DimGrey}{rgb}{0.41,0.41,0.41}
\definecolor{DodgerBlue1}{rgb}{0.12,0.56,1.00}
\definecolor{DodgerBlue2}{rgb}{0.11,0.53,0.93}
\definecolor{DodgerBlue3}{rgb}{0.09,0.45,0.80}
\definecolor{DodgerBlue4}{rgb}{0.06,0.31,0.55}
\definecolor{DodgerBlue}{rgb}{0.12,0.56,1.00}
\definecolor{FloralWhite}{rgb}{1.00,0.98,0.94}
\definecolor{ForestGreen}{rgb}{0.13,0.55,0.13}
\definecolor{GhostWhite}{rgb}{0.97,0.97,1.00}
\definecolor{GreenYellow}{rgb}{0.68,1.00,0.18}
\definecolor{HotPink1}{rgb}{1.00,0.43,0.71}
\definecolor{HotPink2}{rgb}{0.93,0.42,0.65}
\definecolor{HotPink3}{rgb}{0.80,0.38,0.56}
\definecolor{HotPink4}{rgb}{0.55,0.23,0.38}
\definecolor{HotPink}{rgb}{1.00,0.41,0.71}
\definecolor{IndianRed1}{rgb}{1.00,0.42,0.42}
\definecolor{IndianRed2}{rgb}{0.93,0.39,0.39}
\definecolor{IndianRed3}{rgb}{0.80,0.33,0.33}
\definecolor{IndianRed4}{rgb}{0.55,0.23,0.23}
\definecolor{IndianRed}{rgb}{0.80,0.36,0.36}
\definecolor{LavenderBlush1}{rgb}{1.00,0.94,0.96}
\definecolor{LavenderBlush2}{rgb}{0.93,0.88,0.90}
\definecolor{LavenderBlush3}{rgb}{0.80,0.76,0.77}
\definecolor{LavenderBlush4}{rgb}{0.55,0.51,0.53}
\definecolor{LavenderBlush}{rgb}{1.00,0.94,0.96}
\definecolor{LawnGreen}{rgb}{0.49,0.99,0.00}
\definecolor{LemonChiffon1}{rgb}{1.00,0.98,0.80}
\definecolor{LemonChiffon2}{rgb}{0.93,0.91,0.75}
\definecolor{LemonChiffon3}{rgb}{0.80,0.79,0.65}
\definecolor{LemonChiffon4}{rgb}{0.55,0.54,0.44}
\definecolor{LemonChiffon}{rgb}{1.00,0.98,0.80}
\definecolor{LightBlue1}{rgb}{0.75,0.94,1.00}
\definecolor{LightBlue2}{rgb}{0.70,0.87,0.93}
\definecolor{LightBlue3}{rgb}{0.60,0.75,0.80}
\definecolor{LightBlue4}{rgb}{0.41,0.51,0.55}
\definecolor{LightBlue}{rgb}{0.68,0.85,0.90}
\definecolor{LightCoral}{rgb}{0.94,0.50,0.50}
\definecolor{LightCyan1}{rgb}{0.88,1.00,1.00}
\definecolor{LightCyan2}{rgb}{0.82,0.93,0.93}
\definecolor{LightCyan3}{rgb}{0.71,0.80,0.80}
\definecolor{LightCyan4}{rgb}{0.48,0.55,0.55}
\definecolor{LightCyan}{rgb}{0.88,1.00,1.00}
\definecolor{LightGoldenrod1}{rgb}{1.00,0.93,0.55}
\definecolor{LightGoldenrod2}{rgb}{0.93,0.86,0.51}
\definecolor{LightGoldenrod3}{rgb}{0.80,0.75,0.44}
\definecolor{LightGoldenrod4}{rgb}{0.55,0.51,0.30}
\definecolor{LightGoldenrodYellow}{rgb}{0.98,0.98,0.82}
\definecolor{LightGoldenrod}{rgb}{0.93,0.87,0.51}
\definecolor{LightGray}{rgb}{0.83,0.83,0.83}
\definecolor{LightGreen}{rgb}{0.56,0.93,0.56}
\definecolor{LightGrey}{rgb}{0.83,0.83,0.83}
\definecolor{LightPink1}{rgb}{1.00,0.68,0.73}
\definecolor{LightPink2}{rgb}{0.93,0.64,0.68}
\definecolor{LightPink3}{rgb}{0.80,0.55,0.58}
\definecolor{LightPink4}{rgb}{0.55,0.37,0.40}
\definecolor{LightPink}{rgb}{1.00,0.71,0.76}
\definecolor{LightSalmon1}{rgb}{1.00,0.63,0.48}
\definecolor{LightSalmon2}{rgb}{0.93,0.58,0.45}
\definecolor{LightSalmon3}{rgb}{0.80,0.51,0.38}
\definecolor{LightSalmon4}{rgb}{0.55,0.34,0.26}
\definecolor{LightSalmon}{rgb}{1.00,0.63,0.48}
\definecolor{LightSeaGreen}{rgb}{0.13,0.70,0.67}
\definecolor{LightSkyBlue1}{rgb}{0.69,0.89,1.00}
\definecolor{LightSkyBlue2}{rgb}{0.64,0.83,0.93}
\definecolor{LightSkyBlue3}{rgb}{0.55,0.71,0.80}
\definecolor{LightSkyBlue4}{rgb}{0.38,0.48,0.55}
\definecolor{LightSkyBlue}{rgb}{0.53,0.81,0.98}
\definecolor{LightSlateBlue}{rgb}{0.52,0.44,1.00}
\definecolor{LightSlateGray}{rgb}{0.47,0.53,0.60}
\definecolor{LightSlateGrey}{rgb}{0.47,0.53,0.60}
\definecolor{LightSteelBlue1}{rgb}{0.79,0.88,1.00}
\definecolor{LightSteelBlue2}{rgb}{0.74,0.82,0.93}
\definecolor{LightSteelBlue3}{rgb}{0.64,0.71,0.80}
\definecolor{LightSteelBlue4}{rgb}{0.43,0.48,0.55}
\definecolor{LightSteelBlue}{rgb}{0.69,0.77,0.87}
\definecolor{LightYellow1}{rgb}{1.00,1.00,0.88}
\definecolor{LightYellow2}{rgb}{0.93,0.93,0.82}
\definecolor{LightYellow3}{rgb}{0.80,0.80,0.71}
\definecolor{LightYellow4}{rgb}{0.55,0.55,0.48}
\definecolor{LightYellow}{rgb}{1.00,1.00,0.88}
\definecolor{LimeGreen}{rgb}{0.20,0.80,0.20}
\definecolor{MediumAquamarine}{rgb}{0.40,0.80,0.67}
\definecolor{MediumBlue}{rgb}{0.00,0.00,0.80}
\definecolor{MediumOrchid1}{rgb}{0.88,0.40,1.00}
\definecolor{MediumOrchid2}{rgb}{0.82,0.37,0.93}
\definecolor{MediumOrchid3}{rgb}{0.71,0.32,0.80}
\definecolor{MediumOrchid4}{rgb}{0.48,0.22,0.55}
\definecolor{MediumOrchid}{rgb}{0.73,0.33,0.83}
\definecolor{MediumPurple1}{rgb}{0.67,0.51,1.00}
\definecolor{MediumPurple2}{rgb}{0.62,0.47,0.93}
\definecolor{MediumPurple3}{rgb}{0.54,0.41,0.80}
\definecolor{MediumPurple4}{rgb}{0.36,0.28,0.55}
\definecolor{MediumPurple}{rgb}{0.58,0.44,0.86}
\definecolor{MediumSeaGreen}{rgb}{0.24,0.70,0.44}
\definecolor{MediumSlateBlue}{rgb}{0.48,0.41,0.93}
\definecolor{MediumSpringGreen}{rgb}{0.00,0.98,0.60}
\definecolor{MediumTurquoise}{rgb}{0.28,0.82,0.80}
\definecolor{MediumVioletRed}{rgb}{0.78,0.08,0.52}
\definecolor{MidnightBlue}{rgb}{0.10,0.10,0.44}
\definecolor{MintCream}{rgb}{0.96,1.00,0.98}
\definecolor{MistyRose1}{rgb}{1.00,0.89,0.88}
\definecolor{MistyRose2}{rgb}{0.93,0.84,0.82}
\definecolor{MistyRose3}{rgb}{0.80,0.72,0.71}
\definecolor{MistyRose4}{rgb}{0.55,0.49,0.48}
\definecolor{MistyRose}{rgb}{1.00,0.89,0.88}
\definecolor{NavajoWhite1}{rgb}{1.00,0.87,0.68}
\definecolor{NavajoWhite2}{rgb}{0.93,0.81,0.63}
\definecolor{NavajoWhite3}{rgb}{0.80,0.70,0.55}
\definecolor{NavajoWhite4}{rgb}{0.55,0.47,0.37}
\definecolor{NavajoWhite}{rgb}{1.00,0.87,0.68}
\definecolor{NavyBlue}{rgb}{0.00,0.00,0.50}
\definecolor{OldLace}{rgb}{0.99,0.96,0.90}
\definecolor{OliveDrab1}{rgb}{0.75,1.00,0.24}
\definecolor{OliveDrab2}{rgb}{0.70,0.93,0.23}
\definecolor{OliveDrab3}{rgb}{0.60,0.80,0.20}
\definecolor{OliveDrab4}{rgb}{0.41,0.55,0.13}
\definecolor{OliveDrab}{rgb}{0.42,0.56,0.14}
\definecolor{OrangeRed1}{rgb}{1.00,0.27,0.00}
\definecolor{OrangeRed2}{rgb}{0.93,0.25,0.00}
\definecolor{OrangeRed3}{rgb}{0.80,0.22,0.00}
\definecolor{OrangeRed4}{rgb}{0.55,0.15,0.00}
\definecolor{OrangeRed}{rgb}{1.00,0.27,0.00}
\definecolor{PaleGoldenrod}{rgb}{0.93,0.91,0.67}
\definecolor{PaleGreen1}{rgb}{0.60,1.00,0.60}
\definecolor{PaleGreen2}{rgb}{0.56,0.93,0.56}
\definecolor{PaleGreen3}{rgb}{0.49,0.80,0.49}
\definecolor{PaleGreen4}{rgb}{0.33,0.55,0.33}
\definecolor{PaleGreen}{rgb}{0.60,0.98,0.60}
\definecolor{PaleTurquoise1}{rgb}{0.73,1.00,1.00}
\definecolor{PaleTurquoise2}{rgb}{0.68,0.93,0.93}
\definecolor{PaleTurquoise3}{rgb}{0.59,0.80,0.80}
\definecolor{PaleTurquoise4}{rgb}{0.40,0.55,0.55}
\definecolor{PaleTurquoise}{rgb}{0.69,0.93,0.93}
\definecolor{PaleVioletRed1}{rgb}{1.00,0.51,0.67}
\definecolor{PaleVioletRed2}{rgb}{0.93,0.47,0.62}
\definecolor{PaleVioletRed3}{rgb}{0.80,0.41,0.54}
\definecolor{PaleVioletRed4}{rgb}{0.55,0.28,0.36}
\definecolor{PaleVioletRed}{rgb}{0.86,0.44,0.58}
\definecolor{PapayaWhip}{rgb}{1.00,0.94,0.84}
\definecolor{PeachPuff1}{rgb}{1.00,0.85,0.73}
\definecolor{PeachPuff2}{rgb}{0.93,0.80,0.68}
\definecolor{PeachPuff3}{rgb}{0.80,0.69,0.58}
\definecolor{PeachPuff4}{rgb}{0.55,0.47,0.40}
\definecolor{PeachPuff}{rgb}{1.00,0.85,0.73}
\definecolor{PowderBlue}{rgb}{0.69,0.88,0.90}
\definecolor{RosyBrown1}{rgb}{1.00,0.76,0.76}
\definecolor{RosyBrown2}{rgb}{0.93,0.71,0.71}
\definecolor{RosyBrown3}{rgb}{0.80,0.61,0.61}
\definecolor{RosyBrown4}{rgb}{0.55,0.41,0.41}
\definecolor{RosyBrown}{rgb}{0.74,0.56,0.56}
\definecolor{RoyalBlue1}{rgb}{0.28,0.46,1.00}
\definecolor{RoyalBlue2}{rgb}{0.26,0.43,0.93}
\definecolor{RoyalBlue3}{rgb}{0.23,0.37,0.80}
\definecolor{RoyalBlue4}{rgb}{0.15,0.25,0.55}
\definecolor{RoyalBlue}{rgb}{0.25,0.41,0.88}
\definecolor{SaddleBrown}{rgb}{0.55,0.27,0.07}
\definecolor{SandyBrown}{rgb}{0.96,0.64,0.38}
\definecolor{SeaGreen1}{rgb}{0.33,1.00,0.62}
\definecolor{SeaGreen2}{rgb}{0.31,0.93,0.58}
\definecolor{SeaGreen3}{rgb}{0.26,0.80,0.50}
\definecolor{SeaGreen4}{rgb}{0.18,0.55,0.34}
\definecolor{SeaGreen}{rgb}{0.18,0.55,0.34}
\definecolor{SkyBlue1}{rgb}{0.53,0.81,1.00}
\definecolor{SkyBlue2}{rgb}{0.49,0.75,0.93}
\definecolor{SkyBlue3}{rgb}{0.42,0.65,0.80}
\definecolor{SkyBlue4}{rgb}{0.29,0.44,0.55}
\definecolor{SkyBlue}{rgb}{0.53,0.81,0.92}
\definecolor{SlateBlue1}{rgb}{0.51,0.44,1.00}
\definecolor{SlateBlue2}{rgb}{0.48,0.40,0.93}
\definecolor{SlateBlue3}{rgb}{0.41,0.35,0.80}
\definecolor{SlateBlue4}{rgb}{0.28,0.24,0.55}
\definecolor{SlateBlue}{rgb}{0.42,0.35,0.80}
\definecolor{SlateGray1}{rgb}{0.78,0.89,1.00}
\definecolor{SlateGray2}{rgb}{0.73,0.83,0.93}
\definecolor{SlateGray3}{rgb}{0.62,0.71,0.80}
\definecolor{SlateGray4}{rgb}{0.42,0.48,0.55}
\definecolor{SlateGray}{rgb}{0.44,0.50,0.56}
\definecolor{SlateGrey}{rgb}{0.44,0.50,0.56}
\definecolor{SpringGreen1}{rgb}{0.00,1.00,0.50}
\definecolor{SpringGreen2}{rgb}{0.00,0.93,0.46}
\definecolor{SpringGreen3}{rgb}{0.00,0.80,0.40}
\definecolor{SpringGreen4}{rgb}{0.00,0.55,0.27}
\definecolor{SpringGreen}{rgb}{0.00,1.00,0.50}
\definecolor{SteelBlue1}{rgb}{0.39,0.72,1.00}
\definecolor{SteelBlue2}{rgb}{0.36,0.67,0.93}
\definecolor{SteelBlue3}{rgb}{0.31,0.58,0.80}
\definecolor{SteelBlue4}{rgb}{0.21,0.39,0.55}
\definecolor{SteelBlue}{rgb}{0.27,0.51,0.71}
\definecolor{VioletRed1}{rgb}{1.00,0.24,0.59}
\definecolor{VioletRed2}{rgb}{0.93,0.23,0.55}
\definecolor{VioletRed3}{rgb}{0.80,0.20,0.47}
\definecolor{VioletRed4}{rgb}{0.55,0.13,0.32}
\definecolor{VioletRed}{rgb}{0.82,0.13,0.56}
\definecolor{WhiteSmoke}{rgb}{0.96,0.96,0.96}
\definecolor{YellowGreen}{rgb}{0.60,0.80,0.20}
\definecolor{aliceblue}{rgb}{0.94,0.97,1.00}
\definecolor{antiquewhite}{rgb}{0.98,0.92,0.84}
\definecolor{aquamarine1}{rgb}{0.50,1.00,0.83}
\definecolor{aquamarine2}{rgb}{0.46,0.93,0.78}
\definecolor{aquamarine3}{rgb}{0.40,0.80,0.67}
\definecolor{aquamarine4}{rgb}{0.27,0.55,0.45}
\definecolor{aquamarine}{rgb}{0.50,1.00,0.83}
\definecolor{azure1}{rgb}{0.94,1.00,1.00}
\definecolor{azure2}{rgb}{0.88,0.93,0.93}
\definecolor{azure3}{rgb}{0.76,0.80,0.80}
\definecolor{azure4}{rgb}{0.51,0.55,0.55}
\definecolor{azure}{rgb}{0.94,1.00,1.00}
\definecolor{beige}{rgb}{0.96,0.96,0.86}
\definecolor{bisque1}{rgb}{1.00,0.89,0.77}
\definecolor{bisque2}{rgb}{0.93,0.84,0.72}
\definecolor{bisque3}{rgb}{0.80,0.72,0.62}
\definecolor{bisque4}{rgb}{0.55,0.49,0.42}
\definecolor{bisque}{rgb}{1.00,0.89,0.77}
\definecolor{black}{rgb}{0.00,0.00,0.00}
\definecolor{blanchedalmond}{rgb}{1.00,0.92,0.80}
\definecolor{blue1}{rgb}{0.00,0.00,1.00}
\definecolor{blue2}{rgb}{0.00,0.00,0.93}
\definecolor{blue3}{rgb}{0.00,0.00,0.80}
\definecolor{blue4}{rgb}{0.00,0.00,0.55}
\definecolor{blueviolet}{rgb}{0.54,0.17,0.89}
\definecolor{blue}{rgb}{0.00,0.00,1.00}
\definecolor{brown1}{rgb}{1.00,0.25,0.25}
\definecolor{brown2}{rgb}{0.93,0.23,0.23}
\definecolor{brown3}{rgb}{0.80,0.20,0.20}
\definecolor{brown4}{rgb}{0.55,0.14,0.14}
\definecolor{brown}{rgb}{0.65,0.16,0.16}
\definecolor{burlywood1}{rgb}{1.00,0.83,0.61}
\definecolor{burlywood2}{rgb}{0.93,0.77,0.57}
\definecolor{burlywood3}{rgb}{0.80,0.67,0.49}
\definecolor{burlywood4}{rgb}{0.55,0.45,0.33}
\definecolor{burlywood}{rgb}{0.87,0.72,0.53}
\definecolor{cadetblue}{rgb}{0.37,0.62,0.63}
\definecolor{chartreuse1}{rgb}{0.50,1.00,0.00}
\definecolor{chartreuse2}{rgb}{0.46,0.93,0.00}
\definecolor{chartreuse3}{rgb}{0.40,0.80,0.00}
\definecolor{chartreuse4}{rgb}{0.27,0.55,0.00}
\definecolor{chartreuse}{rgb}{0.50,1.00,0.00}
\definecolor{chocolate1}{rgb}{1.00,0.50,0.14}
\definecolor{chocolate2}{rgb}{0.93,0.46,0.13}
\definecolor{chocolate3}{rgb}{0.80,0.40,0.11}
\definecolor{chocolate4}{rgb}{0.55,0.27,0.07}
\definecolor{chocolate}{rgb}{0.82,0.41,0.12}
\definecolor{coral1}{rgb}{1.00,0.45,0.34}
\definecolor{coral2}{rgb}{0.93,0.42,0.31}
\definecolor{coral3}{rgb}{0.80,0.36,0.27}
\definecolor{coral4}{rgb}{0.55,0.24,0.18}
\definecolor{coral}{rgb}{1.00,0.50,0.31}
\definecolor{cornflowerblue}{rgb}{0.39,0.58,0.93}
\definecolor{cornsilk1}{rgb}{1.00,0.97,0.86}
\definecolor{cornsilk2}{rgb}{0.93,0.91,0.80}
\definecolor{cornsilk3}{rgb}{0.80,0.78,0.69}
\definecolor{cornsilk4}{rgb}{0.55,0.53,0.47}
\definecolor{cornsilk}{rgb}{1.00,0.97,0.86}
\definecolor{cyan1}{rgb}{0.00,1.00,1.00}
\definecolor{cyan2}{rgb}{0.00,0.93,0.93}
\definecolor{cyan3}{rgb}{0.00,0.80,0.80}
\definecolor{cyan4}{rgb}{0.00,0.55,0.55}
\definecolor{cyan}{rgb}{0.00,1.00,1.00}
\definecolor{darkblue}{rgb}{0.00,0.00,0.55}
\definecolor{darkcyan}{rgb}{0.00,0.55,0.55}
\definecolor{darkgoldenrod}{rgb}{0.72,0.53,0.04}
\definecolor{darkgray}{rgb}{0.66,0.66,0.66}
\definecolor{darkgreen}{rgb}{0.00,0.39,0.00}
\definecolor{darkgrey}{rgb}{0.66,0.66,0.66}
\definecolor{darkkhaki}{rgb}{0.74,0.72,0.42}
\definecolor{darkmagenta}{rgb}{0.55,0.00,0.55}
\definecolor{darkolive}{rgb}{0.33,0.42,0.18}
\definecolor{darkorange}{rgb}{1.00,0.55,0.00}
\definecolor{darkorchid}{rgb}{0.60,0.20,0.80}
\definecolor{darkred}{rgb}{0.55,0.00,0.00}
\definecolor{darksalmon}{rgb}{0.91,0.59,0.48}
\definecolor{darksea}{rgb}{0.56,0.74,0.56}
\definecolor{darkslate}{rgb}{0.18,0.31,0.31}
\definecolor{darkslate}{rgb}{0.18,0.31,0.31}
\definecolor{darkslate}{rgb}{0.28,0.24,0.55}
\definecolor{darkturquoise}{rgb}{0.00,0.81,0.82}
\definecolor{darkviolet}{rgb}{0.58,0.00,0.83}
\definecolor{deeppink}{rgb}{1.00,0.08,0.58}
\definecolor{deepsky}{rgb}{0.00,0.75,1.00}
\definecolor{dimgray}{rgb}{0.41,0.41,0.41}
\definecolor{dimgrey}{rgb}{0.41,0.41,0.41}
\definecolor{dodgerblue}{rgb}{0.12,0.56,1.00}
\definecolor{firebrick1}{rgb}{1.00,0.19,0.19}
\definecolor{firebrick2}{rgb}{0.93,0.17,0.17}
\definecolor{firebrick3}{rgb}{0.80,0.15,0.15}
\definecolor{firebrick4}{rgb}{0.55,0.10,0.10}
\definecolor{firebrick}{rgb}{0.70,0.13,0.13}
\definecolor{floralwhite}{rgb}{1.00,0.98,0.94}
\definecolor{forestgreen}{rgb}{0.13,0.55,0.13}
\definecolor{gainsboro}{rgb}{0.86,0.86,0.86}
\definecolor{ghostwhite}{rgb}{0.97,0.97,1.00}
\definecolor{gold1}{rgb}{1.00,0.84,0.00}
\definecolor{gold2}{rgb}{0.93,0.79,0.00}
\definecolor{gold3}{rgb}{0.80,0.68,0.00}
\definecolor{gold4}{rgb}{0.55,0.46,0.00}
\definecolor{goldenrod1}{rgb}{1.00,0.76,0.15}
\definecolor{goldenrod2}{rgb}{0.93,0.71,0.13}
\definecolor{goldenrod3}{rgb}{0.80,0.61,0.11}
\definecolor{goldenrod4}{rgb}{0.55,0.41,0.08}
\definecolor{goldenrod}{rgb}{0.85,0.65,0.13}
\definecolor{gold}{rgb}{1.00,0.84,0.00}
\definecolor{gray0}{rgb}{0.00,0.00,0.00}
\definecolor{gray100}{rgb}{1.00,1.00,1.00}
\definecolor{gray10}{rgb}{0.10,0.10,0.10}
\definecolor{gray11}{rgb}{0.11,0.11,0.11}
\definecolor{gray12}{rgb}{0.12,0.12,0.12}
\definecolor{gray13}{rgb}{0.13,0.13,0.13}
\definecolor{gray14}{rgb}{0.14,0.14,0.14}
\definecolor{gray15}{rgb}{0.15,0.15,0.15}
\definecolor{gray16}{rgb}{0.16,0.16,0.16}
\definecolor{gray17}{rgb}{0.17,0.17,0.17}
\definecolor{gray18}{rgb}{0.18,0.18,0.18}
\definecolor{gray19}{rgb}{0.19,0.19,0.19}
\definecolor{gray1}{rgb}{0.01,0.01,0.01}
\definecolor{gray20}{rgb}{0.20,0.20,0.20}
\definecolor{gray21}{rgb}{0.21,0.21,0.21}
\definecolor{gray22}{rgb}{0.22,0.22,0.22}
\definecolor{gray23}{rgb}{0.23,0.23,0.23}
\definecolor{gray24}{rgb}{0.24,0.24,0.24}
\definecolor{gray25}{rgb}{0.25,0.25,0.25}
\definecolor{gray26}{rgb}{0.26,0.26,0.26}
\definecolor{gray27}{rgb}{0.27,0.27,0.27}
\definecolor{gray28}{rgb}{0.28,0.28,0.28}
\definecolor{gray29}{rgb}{0.29,0.29,0.29}
\definecolor{gray2}{rgb}{0.02,0.02,0.02}
\definecolor{gray30}{rgb}{0.30,0.30,0.30}
\definecolor{gray31}{rgb}{0.31,0.31,0.31}
\definecolor{gray32}{rgb}{0.32,0.32,0.32}
\definecolor{gray33}{rgb}{0.33,0.33,0.33}
\definecolor{gray34}{rgb}{0.34,0.34,0.34}
\definecolor{gray35}{rgb}{0.35,0.35,0.35}
\definecolor{gray36}{rgb}{0.36,0.36,0.36}
\definecolor{gray37}{rgb}{0.37,0.37,0.37}
\definecolor{gray38}{rgb}{0.38,0.38,0.38}
\definecolor{gray39}{rgb}{0.39,0.39,0.39}
\definecolor{gray3}{rgb}{0.03,0.03,0.03}
\definecolor{gray40}{rgb}{0.40,0.40,0.40}
\definecolor{gray41}{rgb}{0.41,0.41,0.41}
\definecolor{gray42}{rgb}{0.42,0.42,0.42}
\definecolor{gray43}{rgb}{0.43,0.43,0.43}
\definecolor{gray44}{rgb}{0.44,0.44,0.44}
\definecolor{gray45}{rgb}{0.45,0.45,0.45}
\definecolor{gray46}{rgb}{0.46,0.46,0.46}
\definecolor{gray47}{rgb}{0.47,0.47,0.47}
\definecolor{gray48}{rgb}{0.48,0.48,0.48}
\definecolor{gray49}{rgb}{0.49,0.49,0.49}
\definecolor{gray4}{rgb}{0.04,0.04,0.04}
\definecolor{gray50}{rgb}{0.50,0.50,0.50}
\definecolor{gray51}{rgb}{0.51,0.51,0.51}
\definecolor{gray52}{rgb}{0.52,0.52,0.52}
\definecolor{gray53}{rgb}{0.53,0.53,0.53}
\definecolor{gray54}{rgb}{0.54,0.54,0.54}
\definecolor{gray55}{rgb}{0.55,0.55,0.55}
\definecolor{gray56}{rgb}{0.56,0.56,0.56}
\definecolor{gray57}{rgb}{0.57,0.57,0.57}
\definecolor{gray58}{rgb}{0.58,0.58,0.58}
\definecolor{gray59}{rgb}{0.59,0.59,0.59}
\definecolor{gray5}{rgb}{0.05,0.05,0.05}
\definecolor{gray60}{rgb}{0.60,0.60,0.60}
\definecolor{gray61}{rgb}{0.61,0.61,0.61}
\definecolor{gray62}{rgb}{0.62,0.62,0.62}
\definecolor{gray63}{rgb}{0.63,0.63,0.63}
\definecolor{gray64}{rgb}{0.64,0.64,0.64}
\definecolor{gray65}{rgb}{0.65,0.65,0.65}
\definecolor{gray66}{rgb}{0.66,0.66,0.66}
\definecolor{gray67}{rgb}{0.67,0.67,0.67}
\definecolor{gray68}{rgb}{0.68,0.68,0.68}
\definecolor{gray69}{rgb}{0.69,0.69,0.69}
\definecolor{gray6}{rgb}{0.06,0.06,0.06}
\definecolor{gray70}{rgb}{0.70,0.70,0.70}
\definecolor{gray71}{rgb}{0.71,0.71,0.71}
\definecolor{gray72}{rgb}{0.72,0.72,0.72}
\definecolor{gray73}{rgb}{0.73,0.73,0.73}
\definecolor{gray74}{rgb}{0.74,0.74,0.74}
\definecolor{gray75}{rgb}{0.75,0.75,0.75}
\definecolor{gray76}{rgb}{0.76,0.76,0.76}
\definecolor{gray77}{rgb}{0.77,0.77,0.77}
\definecolor{gray78}{rgb}{0.78,0.78,0.78}
\definecolor{gray79}{rgb}{0.79,0.79,0.79}
\definecolor{gray7}{rgb}{0.07,0.07,0.07}
\definecolor{gray80}{rgb}{0.80,0.80,0.80}
\definecolor{gray81}{rgb}{0.81,0.81,0.81}
\definecolor{gray82}{rgb}{0.82,0.82,0.82}
\definecolor{gray83}{rgb}{0.83,0.83,0.83}
\definecolor{gray84}{rgb}{0.84,0.84,0.84}
\definecolor{gray85}{rgb}{0.85,0.85,0.85}
\definecolor{gray86}{rgb}{0.86,0.86,0.86}
\definecolor{gray87}{rgb}{0.87,0.87,0.87}
\definecolor{gray88}{rgb}{0.88,0.88,0.88}
\definecolor{gray89}{rgb}{0.89,0.89,0.89}
\definecolor{gray8}{rgb}{0.08,0.08,0.08}
\definecolor{gray90}{rgb}{0.90,0.90,0.90}
\definecolor{gray91}{rgb}{0.91,0.91,0.91}
\definecolor{gray92}{rgb}{0.92,0.92,0.92}
\definecolor{gray93}{rgb}{0.93,0.93,0.93}
\definecolor{gray94}{rgb}{0.94,0.94,0.94}
\definecolor{gray95}{rgb}{0.95,0.95,0.95}
\definecolor{gray96}{rgb}{0.96,0.96,0.96}
\definecolor{gray97}{rgb}{0.97,0.97,0.97}
\definecolor{gray98}{rgb}{0.98,0.98,0.98}
\definecolor{gray99}{rgb}{0.99,0.99,0.99}
\definecolor{gray9}{rgb}{0.09,0.09,0.09}
\definecolor{gray}{rgb}{0.75,0.75,0.75}
\definecolor{green1}{rgb}{0.00,1.00,0.00}
\definecolor{green2}{rgb}{0.00,0.93,0.00}
\definecolor{green3}{rgb}{0.00,0.80,0.00}
\definecolor{green4}{rgb}{0.00,0.55,0.00}
\definecolor{greenyellow}{rgb}{0.68,1.00,0.18}
\definecolor{green}{rgb}{0.00,1.00,0.00}
\definecolor{grey0}{rgb}{0.00,0.00,0.00}
\definecolor{grey100}{rgb}{1.00,1.00,1.00}
\definecolor{grey10}{rgb}{0.10,0.10,0.10}
\definecolor{grey11}{rgb}{0.11,0.11,0.11}
\definecolor{grey12}{rgb}{0.12,0.12,0.12}
\definecolor{grey13}{rgb}{0.13,0.13,0.13}
\definecolor{grey14}{rgb}{0.14,0.14,0.14}
\definecolor{grey15}{rgb}{0.15,0.15,0.15}
\definecolor{grey16}{rgb}{0.16,0.16,0.16}
\definecolor{grey17}{rgb}{0.17,0.17,0.17}
\definecolor{grey18}{rgb}{0.18,0.18,0.18}
\definecolor{grey19}{rgb}{0.19,0.19,0.19}
\definecolor{grey1}{rgb}{0.01,0.01,0.01}
\definecolor{grey20}{rgb}{0.20,0.20,0.20}
\definecolor{grey21}{rgb}{0.21,0.21,0.21}
\definecolor{grey22}{rgb}{0.22,0.22,0.22}
\definecolor{grey23}{rgb}{0.23,0.23,0.23}
\definecolor{grey24}{rgb}{0.24,0.24,0.24}
\definecolor{grey25}{rgb}{0.25,0.25,0.25}
\definecolor{grey26}{rgb}{0.26,0.26,0.26}
\definecolor{grey27}{rgb}{0.27,0.27,0.27}
\definecolor{grey28}{rgb}{0.28,0.28,0.28}
\definecolor{grey29}{rgb}{0.29,0.29,0.29}
\definecolor{grey2}{rgb}{0.02,0.02,0.02}
\definecolor{grey30}{rgb}{0.30,0.30,0.30}
\definecolor{grey31}{rgb}{0.31,0.31,0.31}
\definecolor{grey32}{rgb}{0.32,0.32,0.32}
\definecolor{grey33}{rgb}{0.33,0.33,0.33}
\definecolor{grey34}{rgb}{0.34,0.34,0.34}
\definecolor{grey35}{rgb}{0.35,0.35,0.35}
\definecolor{grey36}{rgb}{0.36,0.36,0.36}
\definecolor{grey37}{rgb}{0.37,0.37,0.37}
\definecolor{grey38}{rgb}{0.38,0.38,0.38}
\definecolor{grey39}{rgb}{0.39,0.39,0.39}
\definecolor{grey3}{rgb}{0.03,0.03,0.03}
\definecolor{grey40}{rgb}{0.40,0.40,0.40}
\definecolor{grey41}{rgb}{0.41,0.41,0.41}
\definecolor{grey42}{rgb}{0.42,0.42,0.42}
\definecolor{grey43}{rgb}{0.43,0.43,0.43}
\definecolor{grey44}{rgb}{0.44,0.44,0.44}
\definecolor{grey45}{rgb}{0.45,0.45,0.45}
\definecolor{grey46}{rgb}{0.46,0.46,0.46}
\definecolor{grey47}{rgb}{0.47,0.47,0.47}
\definecolor{grey48}{rgb}{0.48,0.48,0.48}
\definecolor{grey49}{rgb}{0.49,0.49,0.49}
\definecolor{grey4}{rgb}{0.04,0.04,0.04}
\definecolor{grey50}{rgb}{0.50,0.50,0.50}
\definecolor{grey51}{rgb}{0.51,0.51,0.51}
\definecolor{grey52}{rgb}{0.52,0.52,0.52}
\definecolor{grey53}{rgb}{0.53,0.53,0.53}
\definecolor{grey54}{rgb}{0.54,0.54,0.54}
\definecolor{grey55}{rgb}{0.55,0.55,0.55}
\definecolor{grey56}{rgb}{0.56,0.56,0.56}
\definecolor{grey57}{rgb}{0.57,0.57,0.57}
\definecolor{grey58}{rgb}{0.58,0.58,0.58}
\definecolor{grey59}{rgb}{0.59,0.59,0.59}
\definecolor{grey5}{rgb}{0.05,0.05,0.05}
\definecolor{grey60}{rgb}{0.60,0.60,0.60}
\definecolor{grey61}{rgb}{0.61,0.61,0.61}
\definecolor{grey62}{rgb}{0.62,0.62,0.62}
\definecolor{grey63}{rgb}{0.63,0.63,0.63}
\definecolor{grey64}{rgb}{0.64,0.64,0.64}
\definecolor{grey65}{rgb}{0.65,0.65,0.65}
\definecolor{grey66}{rgb}{0.66,0.66,0.66}
\definecolor{grey67}{rgb}{0.67,0.67,0.67}
\definecolor{grey68}{rgb}{0.68,0.68,0.68}
\definecolor{grey69}{rgb}{0.69,0.69,0.69}
\definecolor{grey6}{rgb}{0.06,0.06,0.06}
\definecolor{grey70}{rgb}{0.70,0.70,0.70}
\definecolor{grey71}{rgb}{0.71,0.71,0.71}
\definecolor{grey72}{rgb}{0.72,0.72,0.72}
\definecolor{grey73}{rgb}{0.73,0.73,0.73}
\definecolor{grey74}{rgb}{0.74,0.74,0.74}
\definecolor{grey75}{rgb}{0.75,0.75,0.75}
\definecolor{grey76}{rgb}{0.76,0.76,0.76}
\definecolor{grey77}{rgb}{0.77,0.77,0.77}
\definecolor{grey78}{rgb}{0.78,0.78,0.78}
\definecolor{grey79}{rgb}{0.79,0.79,0.79}
\definecolor{grey7}{rgb}{0.07,0.07,0.07}
\definecolor{grey80}{rgb}{0.80,0.80,0.80}
\definecolor{grey81}{rgb}{0.81,0.81,0.81}
\definecolor{grey82}{rgb}{0.82,0.82,0.82}
\definecolor{grey83}{rgb}{0.83,0.83,0.83}
\definecolor{grey84}{rgb}{0.84,0.84,0.84}
\definecolor{grey85}{rgb}{0.85,0.85,0.85}
\definecolor{grey86}{rgb}{0.86,0.86,0.86}
\definecolor{grey87}{rgb}{0.87,0.87,0.87}
\definecolor{grey88}{rgb}{0.88,0.88,0.88}
\definecolor{grey89}{rgb}{0.89,0.89,0.89}
\definecolor{grey8}{rgb}{0.08,0.08,0.08}
\definecolor{grey90}{rgb}{0.90,0.90,0.90}
\definecolor{grey91}{rgb}{0.91,0.91,0.91}
\definecolor{grey92}{rgb}{0.92,0.92,0.92}
\definecolor{grey93}{rgb}{0.93,0.93,0.93}
\definecolor{grey94}{rgb}{0.94,0.94,0.94}
\definecolor{grey95}{rgb}{0.95,0.95,0.95}
\definecolor{grey96}{rgb}{0.96,0.96,0.96}
\definecolor{grey97}{rgb}{0.97,0.97,0.97}
\definecolor{grey98}{rgb}{0.98,0.98,0.98}
\definecolor{grey99}{rgb}{0.99,0.99,0.99}
\definecolor{grey9}{rgb}{0.09,0.09,0.09}
\definecolor{grey}{rgb}{0.75,0.75,0.75}
\definecolor{honeydew1}{rgb}{0.94,1.00,0.94}
\definecolor{honeydew2}{rgb}{0.88,0.93,0.88}
\definecolor{honeydew3}{rgb}{0.76,0.80,0.76}
\definecolor{honeydew4}{rgb}{0.51,0.55,0.51}
\definecolor{honeydew}{rgb}{0.94,1.00,0.94}
\definecolor{hotpink}{rgb}{1.00,0.41,0.71}
\definecolor{indianred}{rgb}{0.80,0.36,0.36}
\definecolor{ivory1}{rgb}{1.00,1.00,0.94}
\definecolor{ivory2}{rgb}{0.93,0.93,0.88}
\definecolor{ivory3}{rgb}{0.80,0.80,0.76}
\definecolor{ivory4}{rgb}{0.55,0.55,0.51}
\definecolor{ivory}{rgb}{1.00,1.00,0.94}
\definecolor{khaki1}{rgb}{1.00,0.96,0.56}
\definecolor{khaki2}{rgb}{0.93,0.90,0.52}
\definecolor{khaki3}{rgb}{0.80,0.78,0.45}
\definecolor{khaki4}{rgb}{0.55,0.53,0.31}
\definecolor{khaki}{rgb}{0.94,0.90,0.55}
\definecolor{lavenderblush}{rgb}{1.00,0.94,0.96}
\definecolor{lavender}{rgb}{0.90,0.90,0.98}
\definecolor{lawngreen}{rgb}{0.49,0.99,0.00}
\definecolor{lemonchiffon}{rgb}{1.00,0.98,0.80}
\definecolor{lightblue}{rgb}{0.68,0.85,0.90}
\definecolor{lightcoral}{rgb}{0.94,0.50,0.50}
\definecolor{lightcyan}{rgb}{0.88,1.00,1.00}
\definecolor{lightgoldenrod}{rgb}{0.93,0.87,0.51}
\definecolor{lightgoldenrod}{rgb}{0.98,0.98,0.82}
\definecolor{lightgray}{rgb}{0.83,0.83,0.83}
\definecolor{lightgreen}{rgb}{0.56,0.93,0.56}
\definecolor{lightgrey}{rgb}{0.83,0.83,0.83}
\definecolor{lightpink}{rgb}{1.00,0.71,0.76}
\definecolor{lightsalmon}{rgb}{1.00,0.63,0.48}
\definecolor{lightsea}{rgb}{0.13,0.70,0.67}
\definecolor{lightsky}{rgb}{0.53,0.81,0.98}
\definecolor{lightslate}{rgb}{0.47,0.53,0.60}
\definecolor{lightslate}{rgb}{0.47,0.53,0.60}
\definecolor{lightslate}{rgb}{0.52,0.44,1.00}
\definecolor{lightsteel}{rgb}{0.69,0.77,0.87}
\definecolor{lightyellow}{rgb}{1.00,1.00,0.88}
\definecolor{limegreen}{rgb}{0.20,0.80,0.20}
\definecolor{linen}{rgb}{0.98,0.94,0.90}
\definecolor{magenta1}{rgb}{1.00,0.00,1.00}
\definecolor{magenta2}{rgb}{0.93,0.00,0.93}
\definecolor{magenta3}{rgb}{0.80,0.00,0.80}
\definecolor{magenta4}{rgb}{0.55,0.00,0.55}
\definecolor{magenta}{rgb}{1.00,0.00,1.00}
\definecolor{maroon1}{rgb}{1.00,0.20,0.70}
\definecolor{maroon2}{rgb}{0.93,0.19,0.65}
\definecolor{maroon3}{rgb}{0.80,0.16,0.56}
\definecolor{maroon4}{rgb}{0.55,0.11,0.38}
\definecolor{maroon}{rgb}{0.69,0.19,0.38}
\definecolor{mediumaquamarine}{rgb}{0.40,0.80,0.67}
\definecolor{mediumblue}{rgb}{0.00,0.00,0.80}
\definecolor{mediumorchid}{rgb}{0.73,0.33,0.83}
\definecolor{mediumpurple}{rgb}{0.58,0.44,0.86}
\definecolor{mediumsea}{rgb}{0.24,0.70,0.44}
\definecolor{mediumslate}{rgb}{0.48,0.41,0.93}
\definecolor{mediumspring}{rgb}{0.00,0.98,0.60}
\definecolor{mediumturquoise}{rgb}{0.28,0.82,0.80}
\definecolor{mediumviolet}{rgb}{0.78,0.08,0.52}
\definecolor{midnightblue}{rgb}{0.10,0.10,0.44}
\definecolor{mintcream}{rgb}{0.96,1.00,0.98}
\definecolor{mistyrose}{rgb}{1.00,0.89,0.88}
\definecolor{moccasin}{rgb}{1.00,0.89,0.71}
\definecolor{navajowhite}{rgb}{1.00,0.87,0.68}
\definecolor{navyblue}{rgb}{0.00,0.00,0.50}
\definecolor{navy}{rgb}{0.00,0.00,0.50}
\definecolor{oldlace}{rgb}{0.99,0.96,0.90}
\definecolor{olivedrab}{rgb}{0.42,0.56,0.14}
\definecolor{orange1}{rgb}{1.00,0.65,0.00}
\definecolor{orange2}{rgb}{0.93,0.60,0.00}
\definecolor{orange3}{rgb}{0.80,0.52,0.00}
\definecolor{orange4}{rgb}{0.55,0.35,0.00}
\definecolor{orangered}{rgb}{1.00,0.27,0.00}
\definecolor{orange}{rgb}{1.00,0.65,0.00}
\definecolor{orchid1}{rgb}{1.00,0.51,0.98}
\definecolor{orchid2}{rgb}{0.93,0.48,0.91}
\definecolor{orchid3}{rgb}{0.80,0.41,0.79}
\definecolor{orchid4}{rgb}{0.55,0.28,0.54}
\definecolor{orchid}{rgb}{0.85,0.44,0.84}
\definecolor{palegoldenrod}{rgb}{0.93,0.91,0.67}
\definecolor{palegreen}{rgb}{0.60,0.98,0.60}
\definecolor{paleturquoise}{rgb}{0.69,0.93,0.93}
\definecolor{paleviolet}{rgb}{0.86,0.44,0.58}
\definecolor{papayawhip}{rgb}{1.00,0.94,0.84}
\definecolor{peachpuff}{rgb}{1.00,0.85,0.73}
\definecolor{peru}{rgb}{0.80,0.52,0.25}
\definecolor{pink1}{rgb}{1.00,0.71,0.77}
\definecolor{pink2}{rgb}{0.93,0.66,0.72}
\definecolor{pink3}{rgb}{0.80,0.57,0.62}
\definecolor{pink4}{rgb}{0.55,0.39,0.42}
\definecolor{pink}{rgb}{1.00,0.75,0.80}
\definecolor{plum1}{rgb}{1.00,0.73,1.00}
\definecolor{plum2}{rgb}{0.93,0.68,0.93}
\definecolor{plum3}{rgb}{0.80,0.59,0.80}
\definecolor{plum4}{rgb}{0.55,0.40,0.55}
\definecolor{plum}{rgb}{0.87,0.63,0.87}
\definecolor{powderblue}{rgb}{0.69,0.88,0.90}
\definecolor{purple1}{rgb}{0.61,0.19,1.00}
\definecolor{purple2}{rgb}{0.57,0.17,0.93}
\definecolor{purple3}{rgb}{0.49,0.15,0.80}
\definecolor{purple4}{rgb}{0.33,0.10,0.55}
\definecolor{purple}{rgb}{0.63,0.13,0.94}
\definecolor{red1}{rgb}{1.00,0.00,0.00}
\definecolor{red2}{rgb}{0.93,0.00,0.00}
\definecolor{red3}{rgb}{0.80,0.00,0.00}
\definecolor{red4}{rgb}{0.55,0.00,0.00}
\definecolor{red}{rgb}{1.00,0.00,0.00}
\definecolor{rosybrown}{rgb}{0.74,0.56,0.56}
\definecolor{royalblue}{rgb}{0.25,0.41,0.88}
\definecolor{saddlebrown}{rgb}{0.55,0.27,0.07}
\definecolor{salmon1}{rgb}{1.00,0.55,0.41}
\definecolor{salmon2}{rgb}{0.93,0.51,0.38}
\definecolor{salmon3}{rgb}{0.80,0.44,0.33}
\definecolor{salmon4}{rgb}{0.55,0.30,0.22}
\definecolor{salmon}{rgb}{0.98,0.50,0.45}
\definecolor{sandybrown}{rgb}{0.96,0.64,0.38}
\definecolor{seagreen}{rgb}{0.18,0.55,0.34}
\definecolor{seashell1}{rgb}{1.00,0.96,0.93}
\definecolor{seashell2}{rgb}{0.93,0.90,0.87}
\definecolor{seashell3}{rgb}{0.80,0.77,0.75}
\definecolor{seashell4}{rgb}{0.55,0.53,0.51}
\definecolor{seashell}{rgb}{1.00,0.96,0.93}
\definecolor{sienna1}{rgb}{1.00,0.51,0.28}
\definecolor{sienna2}{rgb}{0.93,0.47,0.26}
\definecolor{sienna3}{rgb}{0.80,0.41,0.22}
\definecolor{sienna4}{rgb}{0.55,0.28,0.15}
\definecolor{sienna}{rgb}{0.63,0.32,0.18}
\definecolor{skyblue}{rgb}{0.53,0.81,0.92}
\definecolor{slateblue}{rgb}{0.42,0.35,0.80}
\definecolor{slategray}{rgb}{0.44,0.50,0.56}
\definecolor{slategrey}{rgb}{0.44,0.50,0.56}
\definecolor{snow1}{rgb}{1.00,0.98,0.98}
\definecolor{snow2}{rgb}{0.93,0.91,0.91}
\definecolor{snow3}{rgb}{0.80,0.79,0.79}
\definecolor{snow4}{rgb}{0.55,0.54,0.54}
\definecolor{snow}{rgb}{1.00,0.98,0.98}
\definecolor{springgreen}{rgb}{0.00,1.00,0.50}
\definecolor{steelblue}{rgb}{0.27,0.51,0.71}
\definecolor{tan1}{rgb}{1.00,0.65,0.31}
\definecolor{tan2}{rgb}{0.93,0.60,0.29}
\definecolor{tan3}{rgb}{0.80,0.52,0.25}
\definecolor{tan4}{rgb}{0.55,0.35,0.17}
\definecolor{tan}{rgb}{0.82,0.71,0.55}
\definecolor{thistle1}{rgb}{1.00,0.88,1.00}
\definecolor{thistle2}{rgb}{0.93,0.82,0.93}
\definecolor{thistle3}{rgb}{0.80,0.71,0.80}
\definecolor{thistle4}{rgb}{0.55,0.48,0.55}
\definecolor{thistle}{rgb}{0.85,0.75,0.85}
\definecolor{tomato1}{rgb}{1.00,0.39,0.28}
\definecolor{tomato2}{rgb}{0.93,0.36,0.26}
\definecolor{tomato3}{rgb}{0.80,0.31,0.22}
\definecolor{tomato4}{rgb}{0.55,0.21,0.15}
\definecolor{tomato}{rgb}{1.00,0.39,0.28}
\definecolor{turquoise1}{rgb}{0.00,0.96,1.00}
\definecolor{turquoise2}{rgb}{0.00,0.90,0.93}
\definecolor{turquoise3}{rgb}{0.00,0.77,0.80}
\definecolor{turquoise4}{rgb}{0.00,0.53,0.55}
\definecolor{turquoise}{rgb}{0.25,0.88,0.82}
\definecolor{violetred}{rgb}{0.82,0.13,0.56}
\definecolor{violet}{rgb}{0.93,0.51,0.93}
\definecolor{wheat1}{rgb}{1.00,0.91,0.73}
\definecolor{wheat2}{rgb}{0.93,0.85,0.68}
\definecolor{wheat3}{rgb}{0.80,0.73,0.59}
\definecolor{wheat4}{rgb}{0.55,0.49,0.40}
\definecolor{wheat}{rgb}{0.96,0.87,0.70}
\definecolor{whitesmoke}{rgb}{0.96,0.96,0.96}
\definecolor{white}{rgb}{1.00,1.00,1.00}
\definecolor{yellow1}{rgb}{1.00,1.00,0.00}
\definecolor{yellow2}{rgb}{0.93,0.93,0.00}
\definecolor{yellow3}{rgb}{0.80,0.80,0.00}
\definecolor{yellow4}{rgb}{0.55,0.55,0.00}
\definecolor{yellowgreen}{rgb}{0.60,0.80,0.20}
\definecolor{yellow}{rgb}{1.00,1.00,0.00}

\title{Multi-Stage Fault Warning for Large Electric Grids Using Anomaly Detection and Machine Learning}

\author{
  Sanjeev Raja \\
  College of Engineering \\
  University of Michigan\\
  Ann Arbor, MI 42111, USA \\
  \texttt{sanjeevr@umich.edu} \\
   \And
 Ernest Fokou\'e \\
  School of Mathematical Sciences\\
  Rochester Institute of Technology\\
  Rochester, NY 14623, USA \\
  \texttt{epfeqa@rit.edu} \\
}

\begin{document}
\maketitle

\begin{abstract}
In the monitoring of a complex electric grid, it is of paramount importance to provide operators
with early warnings of anomalies detected on the network,  along with a precise classification and diagnosis of the specific  fault type. In this paper, we propose a novel multi-stage early warning system prototype for electric grid fault detection, classification, subgroup discovery, and visualization. In the first stage, a computationally efficient anomaly detection method based on quartiles detects the presence of a
fault in real time. In the second stage, the fault is classified into one of nine pre-defined disaster
scenarios. The time series data are first mapped to highly discriminative features by applying dimensionality reduction based on temporal autocorrelation. The features are then mapped through
one of three classification techniques: support vector machine, random forest, and artificial neural
network. Finally in the third stage, intra-class clustering based on dynamic time warping is used to
characterize the fault with further granularity. Results on the Bonneville Power Administration electric
grid data show that i) the proposed anomaly detector is both fast and accurate; ii) dimensionality
reduction leads to dramatic improvement in classification accuracy and speed; iii) the random forest
method offers the most accurate, consistent, and robust fault classification; and iv) time series
within a given class naturally separate into five distinct clusters which correspond closely to the
geographical distribution of electric grid buses.

\end{abstract}


\section{Introduction} \label{Introduction}
Electric grids are of vital importance in the infrastructure of modern society, serving to ensure the continuous supply of electricity to households and industries. Grid failures lead to significant financial losses for companies and inconvenience for consumers and maintenance personnel. Automatic monitoring of large complex electric grids has thus received considerable attention from the research community in recent years. Many authors have addressed this challenging problem from a variety of angles using a combination of tools from computer science, electrical engineering, statistics, machine learning and artificial intelligence. In this work we present a novel multistage system using statistical anomaly detection and machine learning techniques to detect, classify, and diagnose electrical faults. While our methodology is applicable to any general time series data, we use the Bonneville Power Administration (BPA) dataset to develop and evaluate our techniques. The BPA electric grid comprises 126 distinct stations or buses, and spans the states of Washington and Oregon. Our dataset is represented as a collection of 126 time series, one for each station, and each time series contains 1800 observations corresponding to 60 seconds of data at a sampling rate of 30 Hz. Three sets of observations are available, namely frequency, voltage, and phase angle metrics. 

Stations in a power grid can fail in several distinct ways. Each failure scenario is characterized by unique time-varying frequency, voltage, and phase angle profiles, and entails specialized reparation by operators. We will refer to these scenarios as fault classes. The BPA dataset contains 16 labeled fault classes corresponding to commonly encountered failures. Of these, several were observed to exhibit near identical temporal behavior. To simplify our classification models and avoid overfitting, highly overlapping fault classes were merged. The result is the following nine fault class labels: \textit {Dropped Load, Open AC, Open DC, Open Generator, GMD 2, Ice Storm, McNary Attack, Ponderosa}, and \textit{Quake 1}. Our goal is to automatically identify and communicate the presence and type of fault in real time so that preemptive action can be taken.

\begin{figure}[htp]
\centering
\includegraphics[width=8cm]{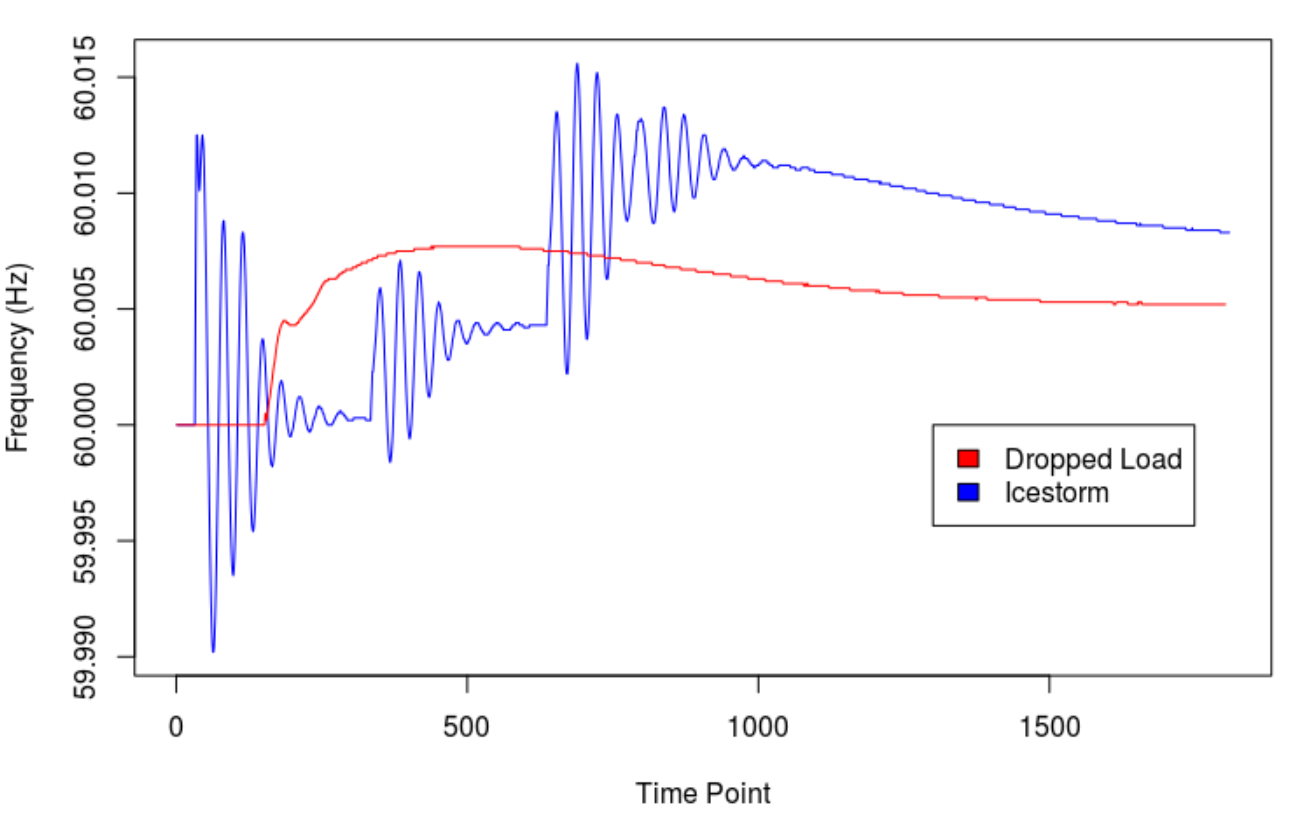}
\caption{Frequency time series of \textit{Ice Storm} and \textit{Dropped Load} faults}
\label{fig:freqplot}
\end{figure}

Figure \ref{fig:freqplot} depicts typical frequency time series for the \textit{Ice Storm} and \textit{Dropped Load} fault classes. The temporal characteristics for these fault types are visibly different.

Recent literature on fault detection and analysis in electric grid systems has largely focused on machine learning approaches \cite{Farshad:2017:1, Hasan:2017:1, Malhotra:2012:1, Tayeb:2013:1, FERREIRA:2018:1, Dalstein:1995:1, Jamil:2015:1, Bhattacharya:2017:1}. Notably, in \cite{FERREIRA:2018:1} a control system approach involving micro-controllers and other electrical equipment was introduced to monitor faults based solely on current fluctuations. In \cite{Dalstein:1995:1} and \cite{Jamil:2015:1}, a fault detection method was combined with classification by an Artificial Neural Network. In \cite{Bhattacharya:2017:1}, a Long-Short Term Memory (LSTM) neural network architecture was proposed to perform fault classification. To our knowledge, investigations exploring random forests or support vector machines, models which are computationally efficient and provide high classification accuracy, are relatively sparse in the literature. Dimensionality reduction and feature selection techniques to reduce computational cost are also not well-developed. Furthermore, subgroup discovery within a fault class using unsupervised learning has not been attempted. Our work is thus novel in that it combines traditional statistical methods such as anomaly detection with advanced machine learning, dimensionality reduction, and visualization, into a cohesive system for real-time fault diagnosis and prediction. 

Prior work by the authors on BPA-specific electric grid data has focused almost exclusively on visualization and exploratory analysis \cite{doi:10.1177/1541931213601976}. This included identifying the Gaussian nature of the distribution of electric frequency measurements across the grid at a specific time point, and generating heat maps of the rate of change of grid parameters across stations. All of the aforementioned analysis was performed for one specific class. In this work, we progress further by analyzing data across multiple classes to develop a warning system that is both highly accurate and computationally efficient, rendering it feasible for expedited implementation at a grid scale.

The rest of this paper is organized as follows: Section \ref{Overview of Early Warning System} provides an overview of the proposed multi-stage fault warning system. Section \ref{Dynamic Anomaly Detection} describes the first stage of the system, namely dynamic anomaly detection. Section \ref{Fault Classification} presents dimensionality reduction and three fault classification methods. Section \ref{Intra-Class Unsupervised Learning} describes intra-class clustering. Experimental results are presented in Section \ref{Experimental Results}, and concluding remarks are given in Section \ref{Conclusions and Future Work}.

\section{Overview of Early Warning System} \label{Overview of Early Warning System}
\begin{figure}[!htbp]
  \centering
  \includegraphics[width=9cm]{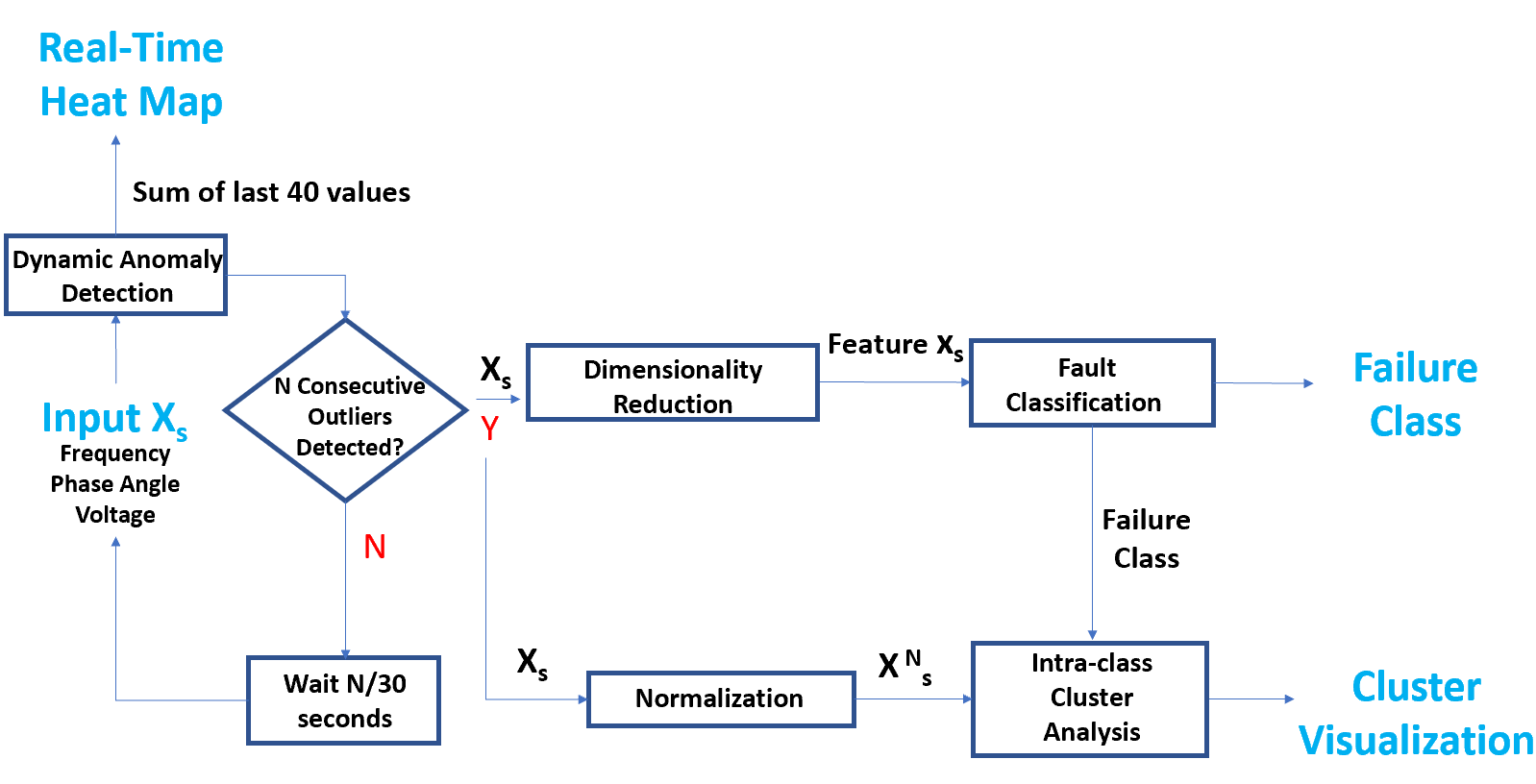}
  \caption{Diagram of proposed early warning system}
  \label{fig:flow}
\end{figure}

We present a framework on which to build a real-time early warning system for BPA electric grid failures, depicted in Figure \ref{fig:flow}. The 126 BPA electric grid stations each produce observations for frequency, voltage, and phase angle. We represent the time series for a station $s$ from time $t=1$ to $t=T$ as a vector $X_{s} = (X_{s,1}, \cdots, X_{s,t}, \cdots, X_{s,T})$, where each $X_{s,t}$ can be a frequency, voltage, or phase angle measurement.The T-dimensional vectors $X_{s}$ are the inputs to the early warning system.

In Stage 1, we perform dynamic anomaly detection on the time series. The results, along with those of the other buses, can be visualized on a heat map. If $n$ consecutive anomalous time points (i.e outliers) are detected in $X_{s}$, where $n$ is an experimentally determined threshold, we assume a fault has occurred, freeze further input to the system from that station, and proceed to Stage 2. Otherwise, we wait $\frac{n}{30}$ seconds (corresponding to a sampling frequency of 30 Hz), receive the latest time series, and conduct Stage 1 analysis again. 
    
In Stage 2, we perform dimensionality reduction on $X_{s}$ to produce a compact representation $x_{s}$ of the time series. We then pass this representation into a classifier (trained offline) to determine the predicted fault class of the time series from a total of 9 possible classes.

In Stage 3, having identified the fault class of $x_{s}$, we perform clustering of data within each class with the aim of gathering insight on the source and severity of a given fault type. 

Each of the stages are described in more detail in the following sections.

\section{Dynamic Anomaly Detection} \label{Dynamic Anomaly Detection}

We employ a simple, robust, and computationally efficient criterion for detecting the presence of anomalous time points or outliers in the electric grid time series data. For a given station s, we map its frequency, voltage, or phase angle time series  $X_{s}$  to an outlier vector $O_{s}$ of the same length. For each time point $X_{s,t}$ in $X_{s}$, we declare the point a moderate outlier if the following condition holds: 

\begin{eqnarray}
X_{s,t} > {\tt Q}_3 + 1.5 \times {\tt IQR} \quad \text{or} \quad X_{s,t} < {\tt Q}_1 - 1.5 \times {\tt IQR}
\label{eq:iqr1}
\end{eqnarray}

where ${\tt Q}_1$ and ${\tt Q}_3$ are the first and third quartile of $X_{s, 1,\cdots,t}$ respectively, and {\tt IQR} is the interquartile range of $X_{s, 1,\cdots,t}$.
If this condition is true, $O_{s,t}$ is set to 1. 
If the following condition holds:
\begin{eqnarray}
X_{s,t} > {\tt Q}_3 + 3 \times {\tt IQR} \quad \text{or} \quad X_{s,t} < {\tt Q}_1 - 3 \times {\tt IQR}
\label{eq:iqr2}
\end{eqnarray}
the time point is considered a severe outlier and the value of $O_{s,t}$  is set to 2. If neither of the above two conditions are true, the point is not considered an outlier, and $O_{s,t}$ is set to 0. 

We thus generate in real time a ternary outlier vector $O_{s,t}$ describing anomaly state as the frequency, voltage, or phase angle measurements from a bus are received. Only time series that exhibit a number of consecutive anomalous
observations are passed to the fault classification stage.

\section{Fault Classification} \label{Fault Classification}

     We use supervised learning techniques to classify the electric grid time series into one of the nine labeled fault classes. As a first step we extract compact features from the raw electrical time series via a dimensionality reduction step, described next. 
     

\subsection{Dimensionality Reduction} \label{Dimensionality Reduction}
A crucial aspect of any classification task is identifying discriminative features of the data on which to perform classification. Raw data often disguises these features and contains considerable redundancy, reducing the accuracy and efficiency of classification. We opt to use the autocorrelation (ACF) function to produce a time-invariant, compact feature representation of the data. ACF returns a vector of values between -1 and 1 representing the correlation of a time series with lagged copies of itself.  Note that the ACF output has constant, specified dimensionality regardless of the dimensionality of the input.

Since ACF is only meaningful on stationary time series, we conduct first-order differencing of the time series $X_{s}$ to ensure stationarity. The differenced time series $\hat{X}_{s}$ is given by:
\begin{eqnarray}
\hat{X}_{s,t} = X_{s,t}-X_{s,t-1}, for \   t=2,\cdots,T,  with\   \hat{X}_{s,1} = 0
\label{eq:diff}
\end{eqnarray}

The autocovariance function at lag $h$ is:
\begin{eqnarray}
          \gamma_s(h) &=& \mathsf{cov}(\hat{X}_{s, t+h},\hat{X}_{s,t}) \nonumber \\
          &=& \bbE\lrsb{(\hat{X}_{s, t+h}-\mu_s)(\hat{X}_{s,t}-\mu_s)}.
          \label{eq:acf1}
        \end{eqnarray}
The corresponding autocorrelation function is:
\begin{eqnarray}
          \rho_s(h) &=& \frac{\gamma_s(h)}{\gamma_s(0)} =  \mathsf{corr}(\hat{X}_{s, t+h},\hat{X}_{s,t})
          \label{eq:acf2}
\end{eqnarray}

For electrical data from a given substation, and for a fixed lag $h$, we obtain a scalar autocorrelation value. By computing this value for $K$ different lags, we obtain a $K$-dimensional feature vector denoted $x_s$ which is then used as input to the classifiers described next.

\subsection{Classification Techniques}
The input to the classifier is feature $x_{s}$ ∈ ${\rm I\!R}$ $^{K}$, and the output is a label $y_{s}$ corresponding to one of nine fault classes. We applied three well known machine learning methods to classify BPA data. We now briefly present these techniques, deferring the reader to the respective references for more detailed descriptions. 

\subsubsection{Support Vector Machine}
A Support Vector Machine (SVM) is a supervised learning method for data classification tasks \cite{Vapnik:97:1}. Given a set of training data with class labels, SVM learns hyperplanes (i.e linear decision boundaries) that optimally divide the data by class. A test point is then assigned to a class based on which side of the hyperplane it falls into. SVM handles more complex classification problems with nonlinear decision surfaces by first transforming input vectors through a nonlinear mapping to a very high-dimensional feature space prior and constructing linear hyperplanes in this space. The derivation of the hyperplane is outside the scope of this paper (see \cite{Vapnik:97:1}); we simply note that the parameters of the hyperplane depend only on a few support vectors, $x_{sj}$ which are in effect the data points closest to the decision boundary.  SVM is well-suited to classification of high-dimensional, sparse data, and is fast to execute, making it an apt technique for classifying BPA electric grid data.

For illustration consider a 2-class problem with inputs $x_s$ and output class labels +1 and -1. The SVM classifier is expressed by the following equation:

\begin{eqnarray}
\hat{f}(\vx_{s}) = {\rm sign}\left(\sum_{j=1}^{N}{\hat{\alpha}_{s_j}\ry_{s_j}K(\vx_{s_j},\vx_{s}) + \hat{b}}\right)
\label{eq:svm}
\end{eqnarray}
where $x_{s_j}$ are $N$ support vectors, $K()$ is a kernel function that takes the input samples into a high-dimensional space, and $\hat{\alpha}_{s_j}$ and $\hat{b}$ are the parameters of the separating hyperplane learned during training. The binary classification problem can be readily extended to multiple classes \cite{Vapnik:97:1}.



\subsubsection{Random Forest}
\begin{figure}[!htbp]
  \centering
  \includegraphics[width=9cm]{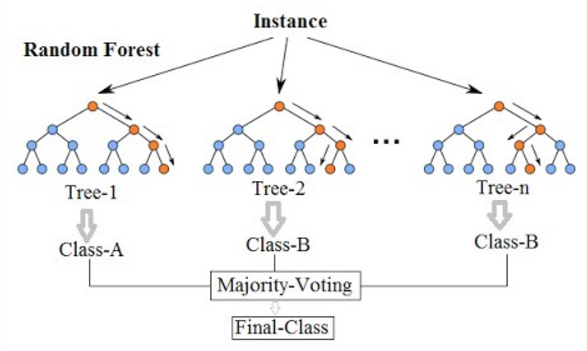}
  \caption{Conceptual illustration of random forest classifier (figure courtesy of \cite{rfpic})}
  \label{fig:rf}
\end{figure}

Random Forests (RF) enable a probabilistic ensemble learning method for data classification \cite{Breiman:2001:1}. The basic block of an RF is a decision tree which recursively splits the $K$-dimensional feature space until a partition of $P$ classes is produced (in our application, $P=9$).  In the case of a binary decision tree, each node of the tree splits the K-dimensional space into two partitions. Repeated splits are performed until the $P$-sized partition is achieved. The parameters of the splits are optimized during a training phase to minimize an overall classification error. One of the shortcomings of decision trees is that they tend to overfit on training data. To mitigate this issue, the RF algorithm aggregates decisions from multiple decision trees. Essentially an RF is comprised a forest of trees, whereby each tree is learned from a random sample of the training data, and the optimal split at each node of a tree is chosen from a random sample of the features of the training data. The output of the RF classifier is the mode (i.e. majority vote) of the class labels predicted by the individual trees, along with a probability of the predicted class membership. A schematic of the RF algorithm is shown in Fig \ref{fig:rf}. Due to the feature diversity offered by a large collection of trees, RF classifiers are generally robust to overfitting, an important consideration for a task such as electrical fault classification with potentially complex decision boundaries.

\subsubsection{Artificial Neural Network}
\begin{figure}[htp]
\centering
\includegraphics[width=8cm]{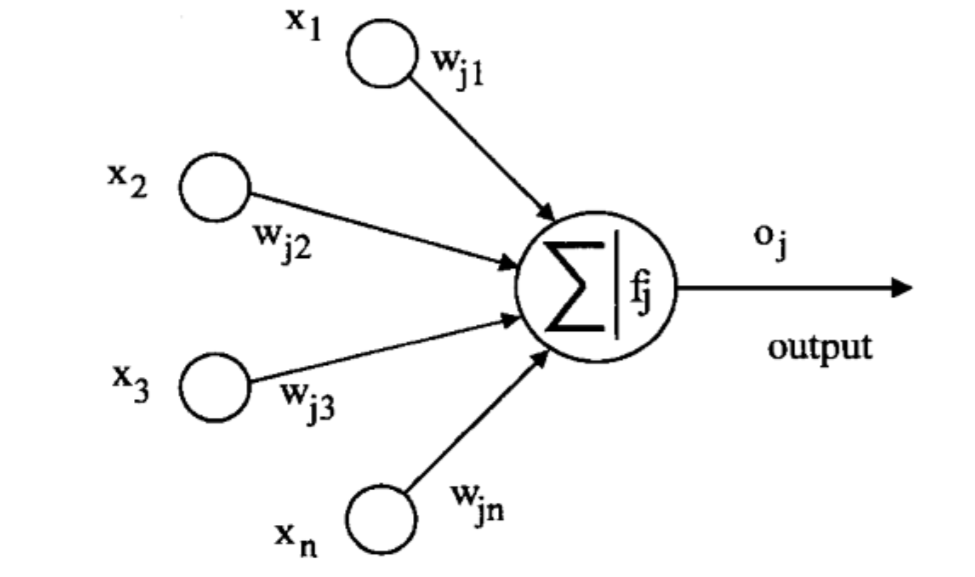}
\caption{A depiction of a typical node in an Artificial Neural Network, where $x_i$ denotes input, $o_j$ denotes output, $w$ denotes connection weights, and $f_j$ is a nonlinear activation function (image courtesy of \cite{Dalstein:1995:1})}.
\label{fig:node}
\end{figure}

An artificial neural network (ANN) is a group of interconnected nodes organized into layers: an input layer with one node for each input feature, several hidden layers, and an output layer with one node for each possible output \cite{BishopCM}. Every inter-node connection in the network carries an associated weight and a function that maps an input to a known output. This is illustrated for a single node in Figure \ref{fig:node}.

The computations at the single node of Fig 3 are given by: 
\begin{eqnarray}
\eta(\vx) = o_j = f_j(\beta_{j}+ \sum_{i}{w_{ji} x_{i}})
\label{eq:ann}
\end{eqnarray}
where $\beta_{j}$ is a learnable scalar bias term.
An ANN combines multiple nodes in multiple cascaded layers to realize arbitrarily complex classification and regression functions.  Several choices exist for the activation function, such as the logistic sigmoid function:
\begin{eqnarray}
\psi(z) = \frac{1}{1+e^{-z}}
\label{eq:act1}
\end{eqnarray}

and hyperbolic tangent function:
\begin{eqnarray}
\psi(z) = {\rm tanh}(z) =  \frac{e^{z}-e^{-z}}{e^{z}+e^{-z}} = \frac{1-e^{-2z}}{1+e^{-2z}}
\label{eq:act2}
\end{eqnarray}

We choose the hyperbolic tangent function due to its favorable convergence properties.

During training, the network weights $w_{ji}$ are first initialized, usually with random coefficients. The network processes training samples one at a time, comparing the predicted output of the network to the ground truth. This comparison yields an error that is back-propagated through the system, and the node connection weights $w_{ji}$ are adjusted to minimize training error. This process iterates until convergence (see \cite{BishopCM} for a detailed overview of ANN training). ANN classifiers are highly tolerant to noisy data and can learn arbitrarily complex decision boundaries, an important criterion for multinomial classification. In our experiments, we employ a 3-layer ANN, whose details are given in Section \ref{Experimental Results}.

\section {Intra-Class Unsupervised Learning} \label{Intra-Class Unsupervised Learning}
Individual stations in an electric grid can respond differently to a given fault based on geography, size, staffing, and a number of other factors. We use unsupervised learning techniques, namely time series clustering, to discover meaningful subgroups of grid stations within a fault type. These subgroups could potentially correspond to varying severity of a fault or other differentiating factors. With intra-class analysis, grid operators  gain access to more granular information such as specific geographical areas that have been affected most severely and require immediate intervention. While the fault classifier in the previous stage operates on a compact ACF representation $x_s$, in this stage we operate on the original raw time series $X_s$ to uncover temporal structure within each class. 

 Prior to clustering, we normalize the time series so that all values are in the range of [0, 1]. Normalization augments the disparity between times series, allowing for greater clustering acuity.For each $X_{s,t}$ in a time series $X_{s}$, the normalized value $X_{s,t}^{N}$ is given by
 \begin{eqnarray}
X_{s,t}^{N} =\frac{X_{s,t} - \min X_{s}}{\max X_{s} - \min X_{s} } 
 \label{eq:minmaxnorm}
\end{eqnarray}

 \subsection {Dynamic Time Warping}
Any clustering technique requires a metric that defines distance between samples. Since we are dealing with time series, we use dynamic time warping (DTW) \cite{Sakoe:79:1} to compute the similarity (or distance) between two temporal sequences. To briefly review DTW, given two time series $U$ and $V$ of length $m$ and $n$, respectively, an $m$ x $n$ distance matrix $D$ is constructed with elements $D_{ij}$ representing the pairwise distance between points $U_i$ and $V_j$ . Euclidean distance is commonly used, such that
\begin{eqnarray}
D_{ij} =  \sqrt{(U_i- V_j)^{2}}
 \label{eq:euc}
\end{eqnarray}

A warping path $w$ is defined as a contiguous sequence of $k$ matrix elements that satisfies the following two conditions:
\begin{enumerate}
    \item Boundary conditions: $w_1$ = $(1,1)$ and $w_k$ = $(m,n)$
    \item Continuity and monotonicity: if $w_i$ = $(a,b)$ then $w_{i-1}$ = $(a^{'}, b^{'})$ where $0$ $\leq$ $a-a^{'}$ $\leq$ $1$ and $0$ $\leq$ $b-b^{'}$ $\leq$ $1$

\end{enumerate}

The cost of the path is defined as the sum of the elements (distances) traversed by the path. DTW seeks the path with minimum cost, the latter being the DTW distance between the two sequences. DTW is often sped up by limiting the set of valid paths to a limited region of matrix $D$ around the diagonal.

\subsection{Partioning Around Medoids}
Given a distance metric, we next proceed to cluster the data using the Partitioning Around Medoids (PAM) algorithm \cite{kaufman:87:1}. PAM is a variant of the well-known K-means clustering algorithm \cite{Lloyd:82:1}. However instead of representing each cluster with its centroid as is done in K-means, PAM represents each cluster with an exemplar data point, referred to as the medoid (or “middle point”). More crucially, PAM admits more general distance metrics, in contrast to K-means which uses only squared Euclidean distance. Hence PAM is favorable for our application where DTW defines distances between time series. 

With the task of partitioning the data in a given fault class into $L$ clusters, PAM proceeds as follows:
\begin{enumerate}
    \item Select $L$ out of the data points as initial medoids
    \item Associate each data point with the closest medoid, with distance measured by DTW
    \item Compute the total cost, which is the sum of distances of points to their assigned medoids
    \item While the total cost decreases:
        \begin {enumerate}
        \item For each medoid $m$ and for each non-medoid sample $d$
        \begin{enumerate}
            \item Swap $m$ and $d$, reassign all points to the closest medoid and compute total cost
            \item If the total cost increased in the previous step,  undo the swap
        \end{enumerate}
        \end {enumerate}
    
\end{enumerate}

\section{Experimental Results} \label{Experimental Results}
We now present results for the BPA data. We used the \textit{R} statistical language to develop all the analysis. Recall that the data contains 126 time series $X_s$, one for each station in the BPA grid. Most time series contained 1802 time samples, corresponding to just over 60 seconds of frequency, voltage, or phase angle data at a sampling rate of 30 Hz, while a few time series were longer, containing up to 3000 samples. In preliminary experiments, we found frequency observations to be more discriminative for fault classification than voltage and phase angle; thus frequency data was used for all experiments. Any further mention of time series refers specifically to frequency time series.

\subsection{Anomaly Detection}

\begin{figure}[ht].
\centering
\includegraphics[width=10 cm]{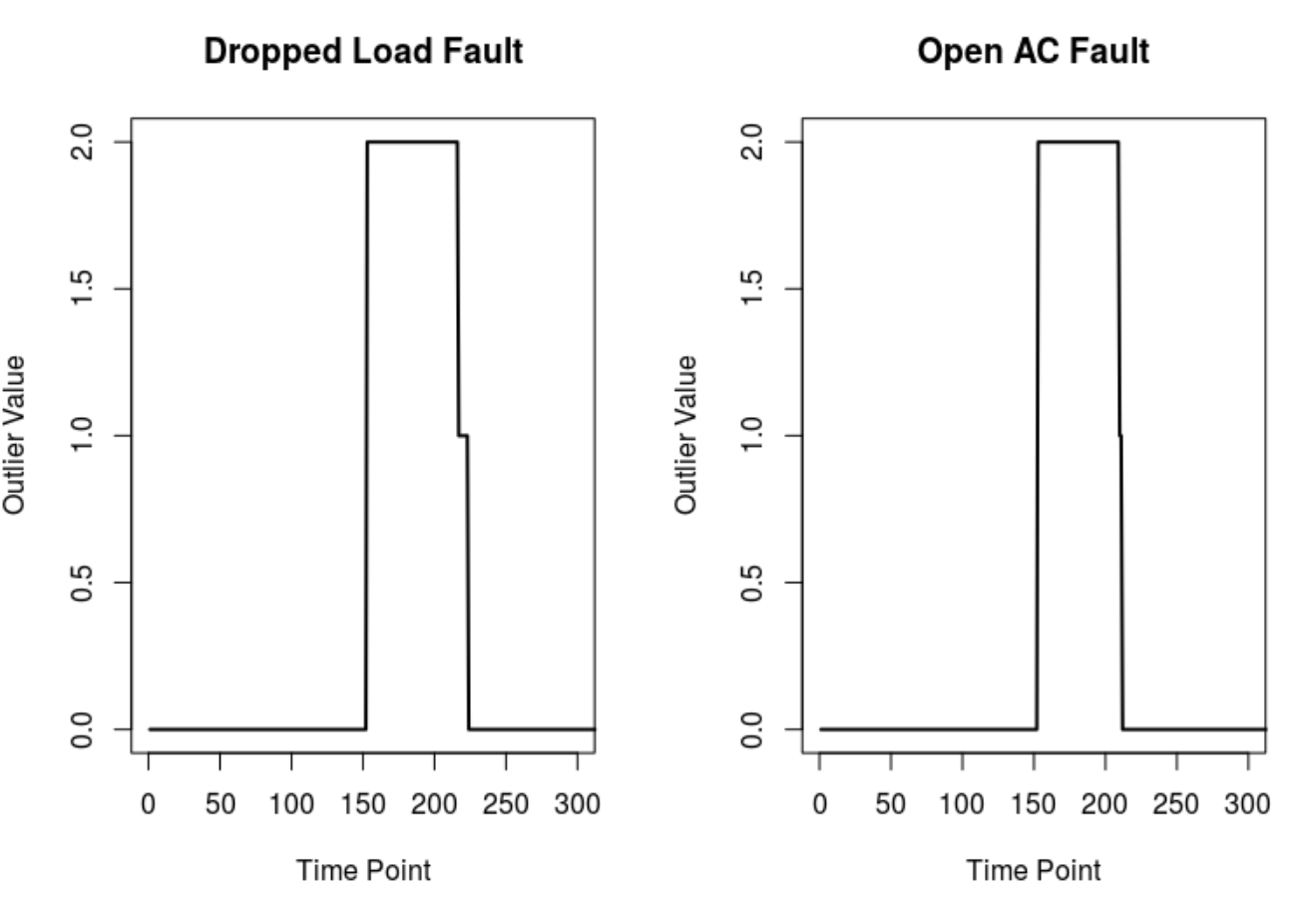}
\caption{Outlier vector for \textit{Dropped Load} and \textit{Open AC} faults generated using quartile criterion. Faults are reliably characterized by a sudden spike of 70 anomalous time points}
\label{fig:spike}
\end{figure}

Applying our quartile criterion to detect outliers in the frequency data, we found that the onset of a fault is characterized by approximately 70 consecutive severe outliers followed by a return to normalcy due to the self-adjusting nature of the outlier detection criterion. This is depicted in Figure \ref{fig:spike}

Based on this signature of fault onset, we propose setting our early warning system outlier threshold at 70. In other words, when we detect 70 consecutive outliers relative to the accepted 60 Hz frequency value, we assume a fault has occurred and proceed to the second stage for more detailed classification. 

\begin{figure}[htp]
\centering
\includegraphics[width=10 cm]{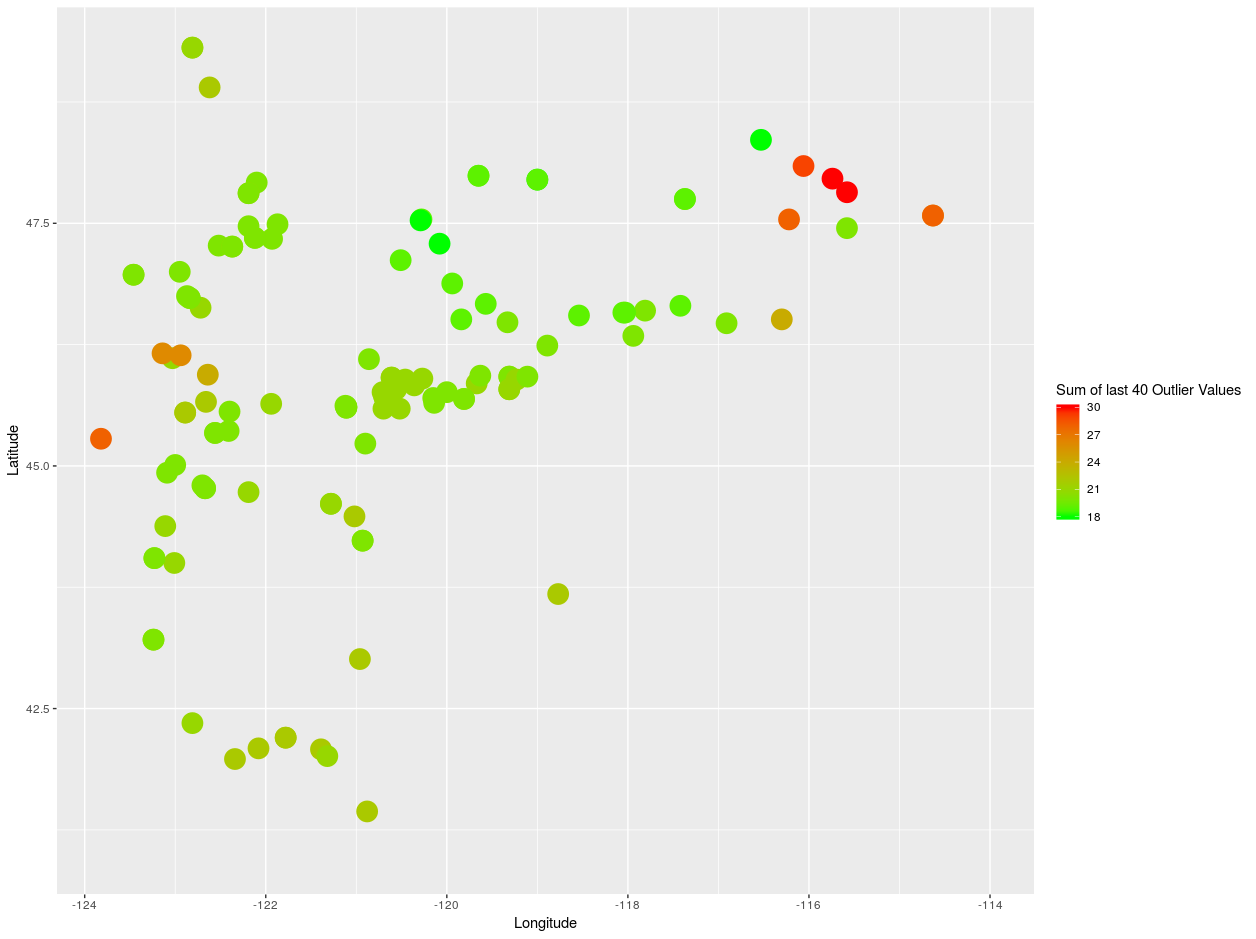}
\caption{Severity of \textit{Icestorm} fault across BPA electric grid stations }
\label{fig:heatmap}
\end{figure}

We also used the outlier vector to generate an intuitive visualization of the BPA electric grid, shown in Figure \ref{fig:heatmap}. The colors of the scatter plot correspond to the sum of the last 40 values of the outlier vector $O_{s}$for a given station (ranging from 0-80). This provides unique insight into the location and severity of the fault, and can be updated in real-time to provide operators a live picture of the health of the grid. 

\subsection {Dimensionality Reduction}
\begin{figure}[htp]
\centering
\includegraphics[width=8cm]{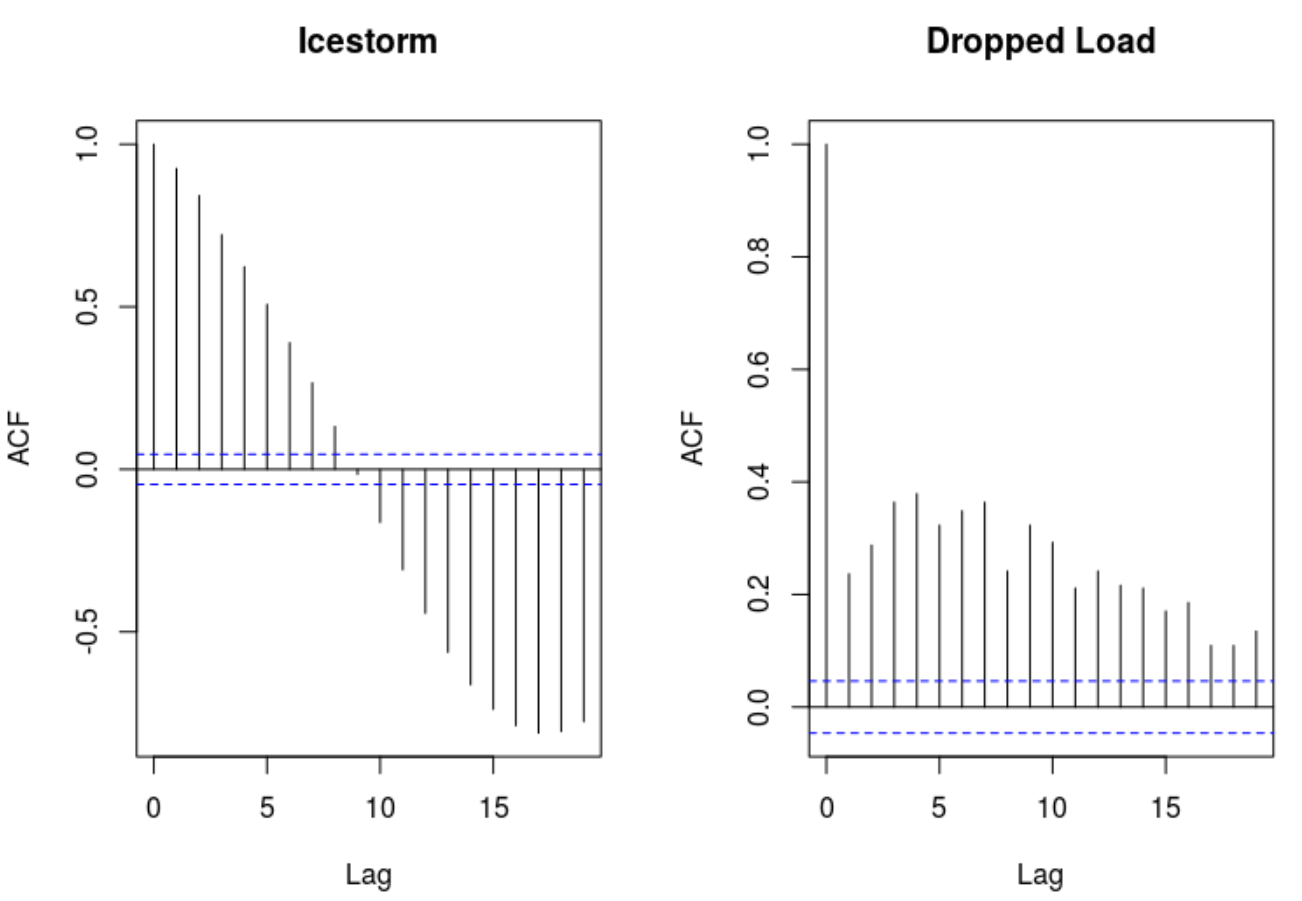}
\caption{Autocorrelation feature vector for differenced frequency series corresponding to \textit{Icestorm} and \textit{Dropped Load} faults}
\label{fig:acf}
\end{figure}

We conducted first-order differencing of the frequency series and applied the ACF function to obtain a compact representation. The maximum number of lags we used was 20. ACF thus condenses our raw data $X_s$ containing 1800 to 3000 time points to a time-invariant 20-dimensional feature vector $x_s$. Figure \ref{fig:acf} shows the ACF feature vector of the differenced frequency series for the \textit{Icestorm} and \textit{Dropped Load} faults.

We performed an ablation study to determine the accuracy of SVM classification with and without dimensionality reduction. Using the original frequency data with 1802 dimensions, it took 21.2 seconds to train an SVM classifier on 1602 examples, and 2.29 seconds to predict the fault class of 399 test time series, with 53.8 percent of test samples being labelled correctly. Using the 20-dimensional ACF representation on the same training and test data, training took 0.25 seconds,  testing took 0.031 seconds, and there was a dramatic increase in classification accuracy to 96.0 percent. Clearly, ACF dimensionality reduction results in a highly discriminative, compact representation, allowing for real-time, accurate classification that cannot be accomplished using raw data. We also experimented with the partial autocorrelation function \cite{Degerine:03:1} and spectral periodogram \cite{Stoica:05:1} obtained via a Fast Fourier Transform and found that both methods were inferior to the ACF in terms of discriminability, classification accuracy, and robustness to inputs of varying dimensions.

\subsection{Fault Classification}
We applied the three classification models introduced above to the task of classifying the BPA time series into nine fault scenarios. 

\subsubsection{Support Vector Machine}
We used the radial basis function with a gamma value of 0.05 as our SVM kernel. We set our soft margin cost parameter to 1. The two fault classes that were a result of consolidation contained approximately 500 data examples each, while the others contained 126 examples each corresponding to the 126 buses in the BPA electric grid. Out of a total of 2001 data examples, 80 percent (1602) were used to train the classifier, and the remaining 20 percent (399) were reserved for testing. The training input to the classifier was a set of 20-dimensional ACF vectors of first-order differenced time series along with corresponding fault class labels. At inference, the input to the classifier was an ACF vector, and the output was the predicted fault class. The mean classification accuracy of the SVM classifier on the testing data, calculated over 100 trials on different training-testing splits, was 97.2 percent.

\subsubsection{Random Forest}

We implemented an RF classifier with 500 trees, determined to be optimal from hyperparameter tuning. The same protocols used for training and evaluating the SVM classifier were also used for the RF classifier. The mean classification accuracy of the RF classifier on the testing data, calculated over 100 trials on different training-testing splits, was 98.9 percent.

\subsubsection{Artificial Neural Network}
\begin{figure}[htp]
\centering
\includegraphics[width=8cm]{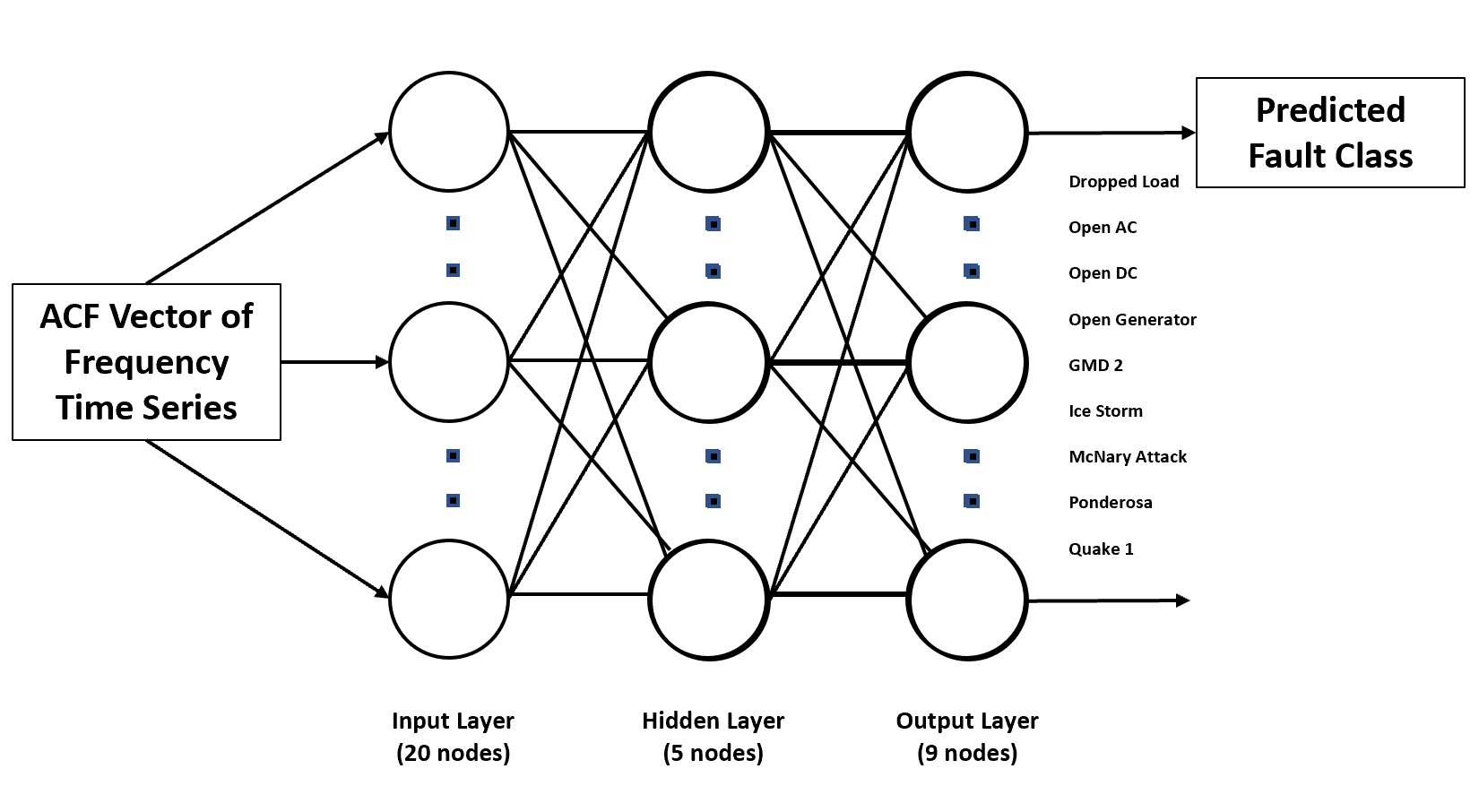}
\caption{Artificial neural network for fault classification}
\label{fig:ann}
\end{figure}

We implemented the ANN classifier depicted in Figure \ref{fig:ann}. The network consists of an input layer of 20 nodes corresponding to the 20-dimensional ACF vector, a single hidden layer of 5 nodes, and an output layer of 9 nodes for each target fault class. The same protocols used for the previous classifiers were used to train and evaluate the ANN. The mean classification accuracy calculated over 100 trials on different training-testing splits was 97.8 percent.

\subsubsection{Comparison of Classifiers}
\begin{figure}[htp]
\centering
\includegraphics[width=8cm]{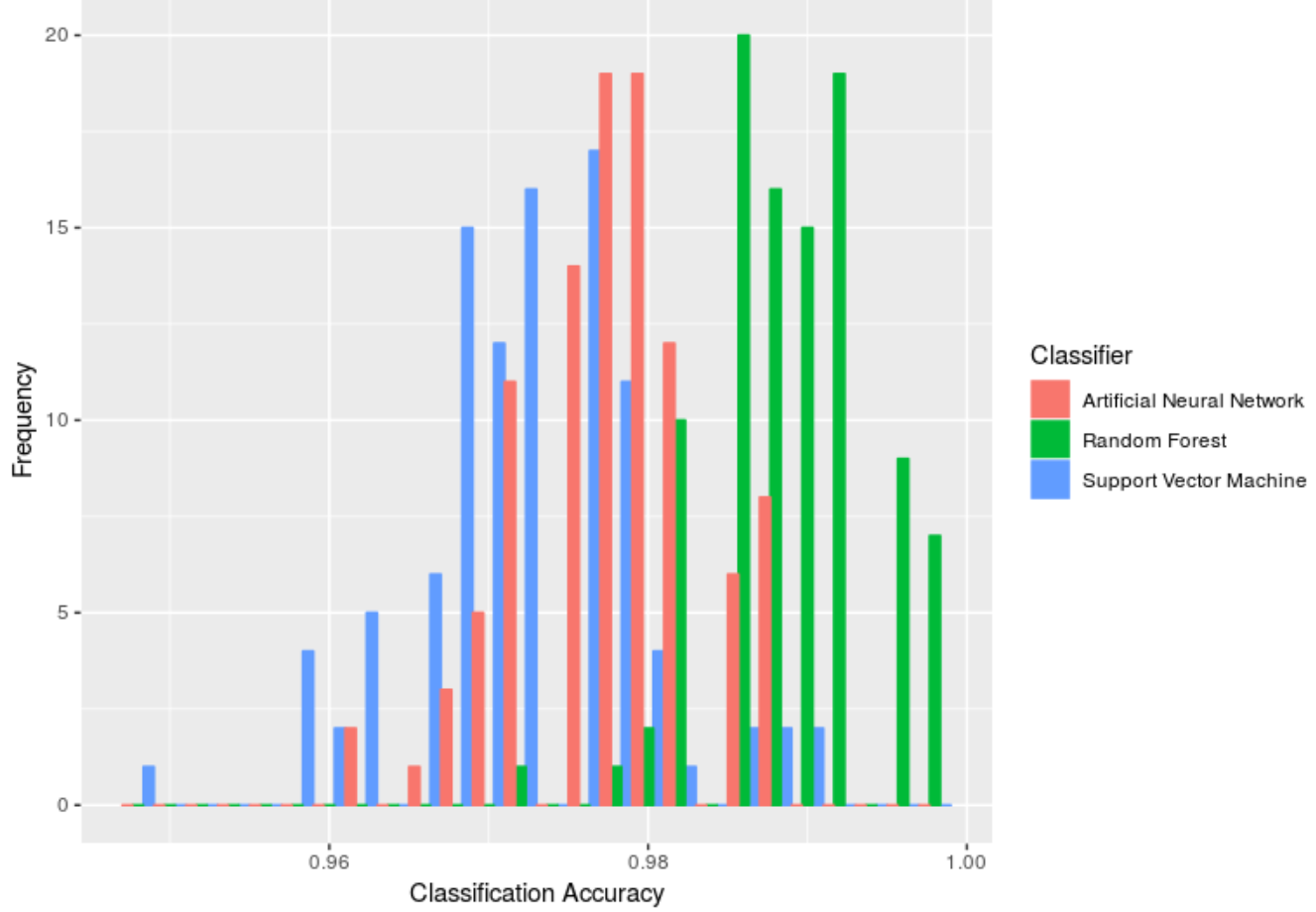}
\caption{Histogram of RF, SVM, and ANN classifier accuracy over 100 trials (accuracy=1 corresponds to 100 percent of test samples being classified correctly)}
\label{fig:hist}
\end{figure}
Figure \ref{fig:hist} compares histograms of accuracy scores for the three classifiers. Each classifier was run 100 times, with each iteration using a unique train-test split stratified by fault type. We note that the classification accuracy of the RF classifier is consistently higher than that of SVM and ANN. RF classification accuracies are skewed left and are more tightly clustered around their mean of 98.9 percent. SVM classification accuracies are more nearly Gaussian around 97.2 percent with a greater spread, while ANN accuracies are skewed right but with a lower mean of 97.8 percent. We can infer that not only does RF on average outperform SVM and ANN on our classification task, but it also displays less variability and therefore greater reliability across many iterations of classification. 

\begin{figure}[htp]
\centering
\includegraphics[width=8cm]{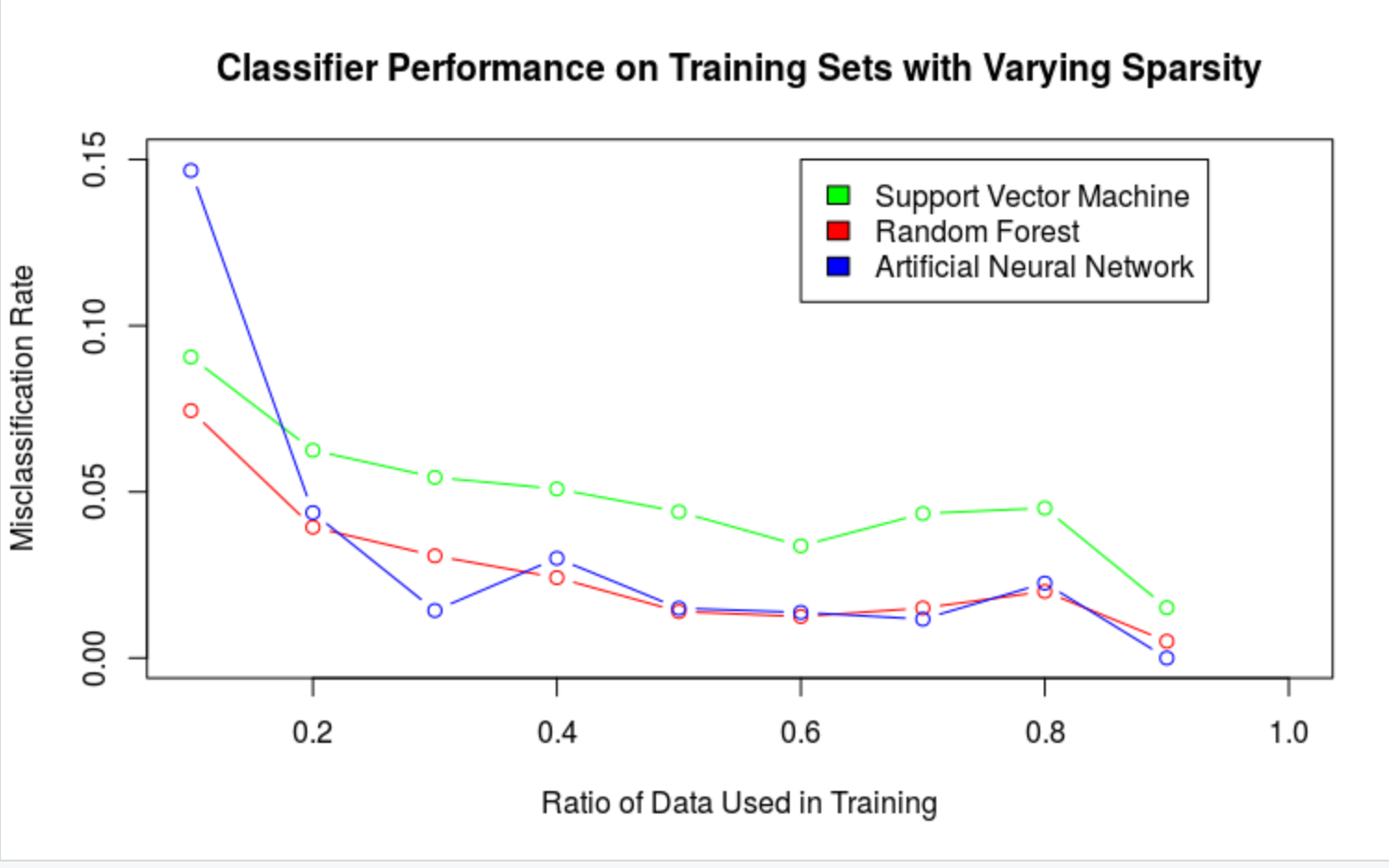}
\caption{Classifier performance as a function of training budget}
\label{fig:sp}
\end{figure}

Figure \ref{fig:sp} depicts the performance of the 3 classifiers as a function of the size of the training set, represented as a fraction of the total number of data samples. It is clear that when at least 20 percent of the data (399 time series) is used in training, both ANN and RF consistently outperform SVM, with the former two being similar in performance. However, when less than 20 percent of the data is used in the training set, ANN’s misclassification rate rises to nearly 15 percent. ANN’s considerable drop in performance can be attributed to its complexity as a model. ANN is able to learn arbitrarily complex decision boundaries, likely causing it to overfit in the presence of sparse training data. SVM and RF are comparatively simpler models that appear more robust to overfitting on this data set. The discrepancy among classifier performances on sparse training sets is highly revealing. For the BPA dataset, both RF and ANN appear to be equally viable candidates for classification, as there are no data limitations. However, when considering the application of our early warning system to other electric grids or domains, availability of training data may be an important consideration. In sparse training scenarios, our results strongly suggest that RF is the premier candidate for fault classification.

\subsubsection{Preemptive Classification}
\begin{figure}[htp]
\centering
\includegraphics[width=8cm]{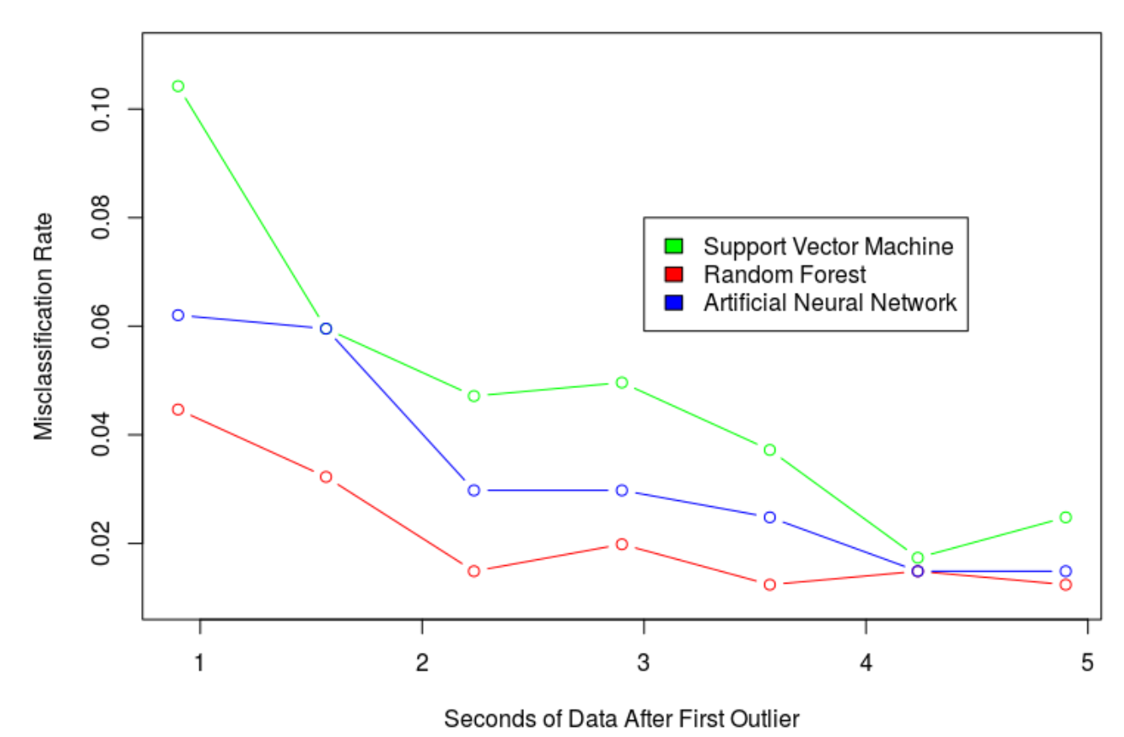}
\caption{Misclassification Rates of RF, SVM, and ANN as a function of number of temporal samples used to define input features}
\label{fig:pre}
\end{figure}

Classification results reported thus far have been on complete time series containing 1802 time points, or approximately 60 seconds of data. Realistically, we aim to classify faults before they have run their entire course. To this end, we couple the outlier detection method with classification. Upon identification of 70 consecutive outliers, we aim to classify the time series as quickly as possible to provide ample time for operators to intervene. We trained the RF, SVM, and ANN classifiers on features derived from varying numbers of time samples captured immediately after the occurrence of the first outlier in the 70-outlier series. ACF was performed on the time series prior to classification. Figure \ref{fig:pre} plots classifier performance as a function of the number of temporal samples used for feature definition.
    
    The results in Figure \ref{fig:pre} bolster the superiority of RF over SVM and ANN in  timely classification of electric grid faults. Notably, RF requires under 2 seconds of additional data beyond the first outlier to identify the fault type of a time series with maximum accuracy. This is a significant enabler for real-time fault prediction.

\subsection { Intra-Class Clustering}
\begin{figure}[htp]
\centering
\includegraphics[width=8cm]{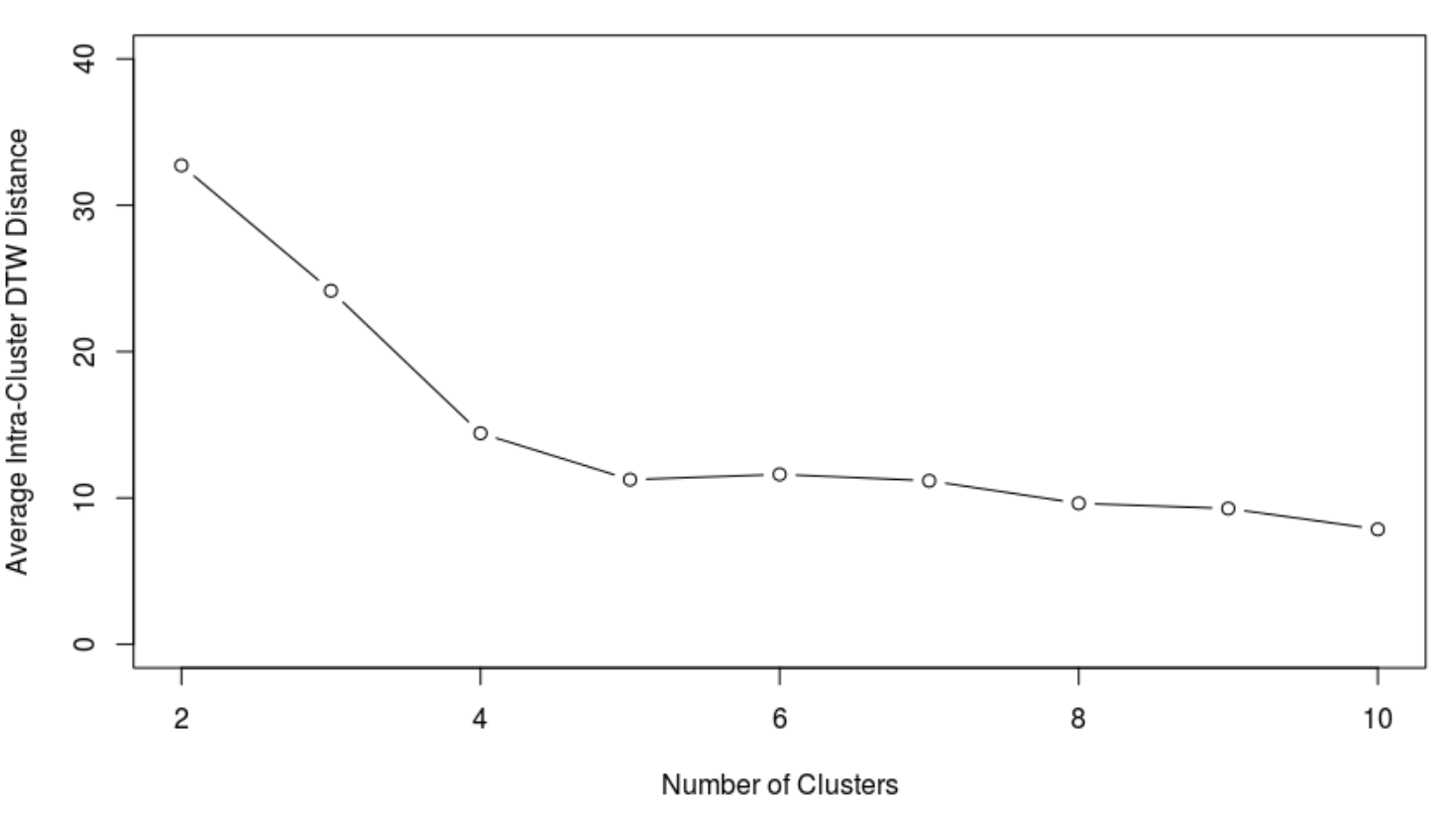}
\caption{Average intra-cluster DTW distance (averaged across all clusters) as a function of the number of clusters }
\label{fig:elbow}
\end{figure}

In this stage, we clustered the data within each fault class to uncover additional structure and insight using DTW as the distance metric and  PAM as the clustering technique. We show results for the \textit{GMD 2} fault class.  The inputs to the clustering were min-max normalized time series. We computed the average intra-cluster DTW distance for different numbers of clusters, plotted in Figure \ref{fig:elbow}. We note from this plot that intra-cluster distance initially drops as the number of clusters increases, and then flattens beyond the case of 5 clusters. We thus determined that the optimal number of clusters to capture structure in the data is 5. The five clusters contained 28, 10, 17, 17, and 54 temporal waveforms, with respective average intra-cluster DTW distances of 17.870670, 5.740929, 14.639842, 17.140088, and 7.272719. 

\begin{figure}[htp]
\centering
\includegraphics[width=8cm]{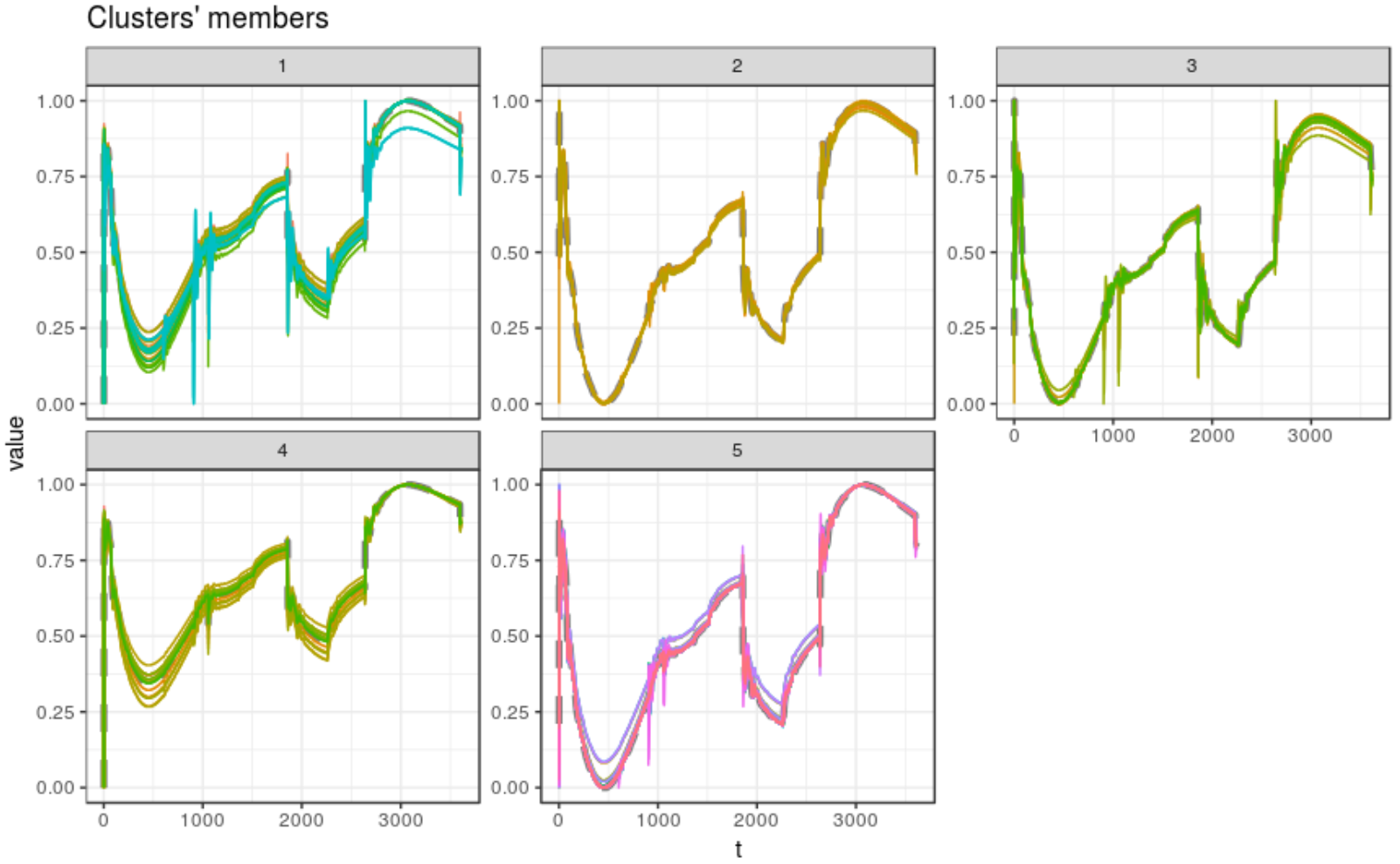}
\caption{Clustering of GMD 2 fault time series (time series are min-max normalized)}
\label{fig:clusts}
\end{figure}

The five clusters contained 28, 10, 17, 17, and 54 time series, with an average intra-cluster DTW distance of  17.870670, 5.740929, 14.639842, 17.140088, and 7.272719 respectively. Figure \ref{fig:clusts} shows the normalized time series in each cluster.

The time series appear to be sufficiently different across clusters. Although there is some intra-cluster varation,  splitting the data into additional clusters may cause overfitting.

\begin{figure}[htp]
\centering
\includegraphics[width=8cm]{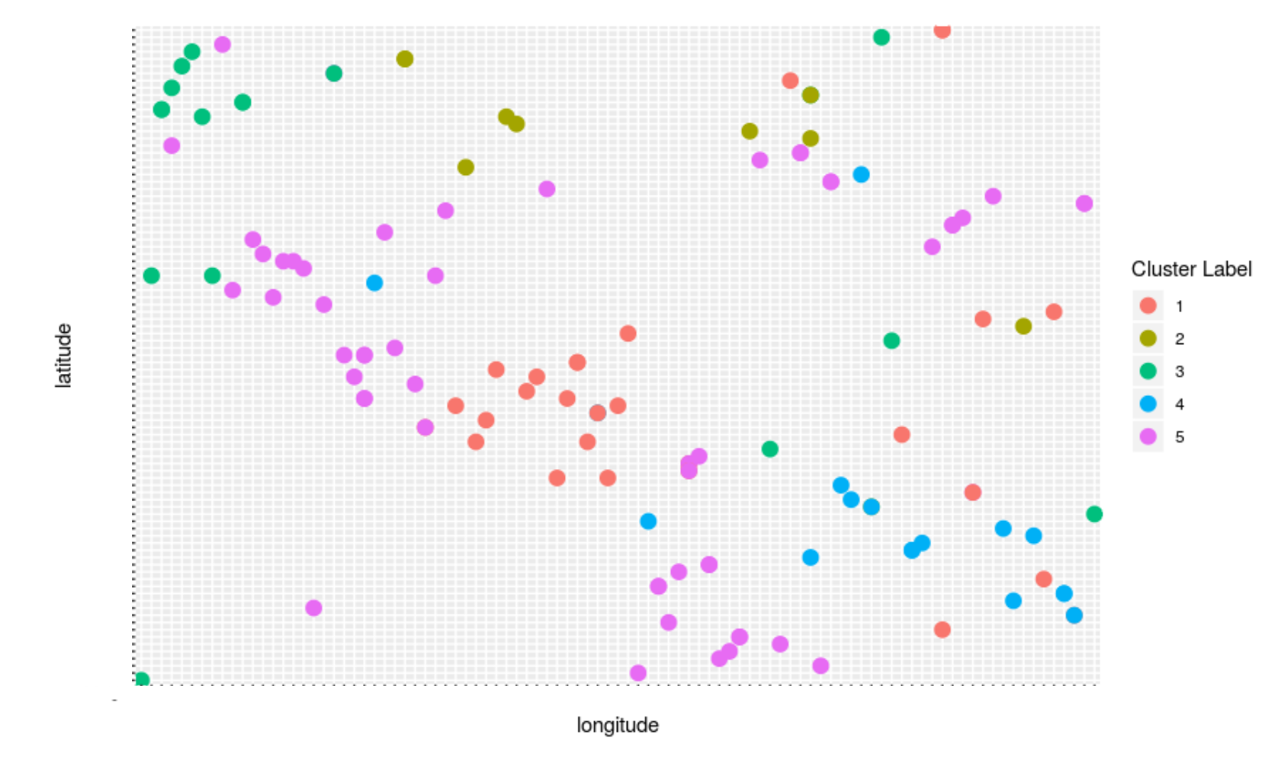}
\caption{Geographical representation of clustering of GMD 2 fault time series}
\label{fig:geo}
\end{figure}

Figure \ref{fig:geo} plots the geographical coordinates of the GMD 2 faults, color-coded by cluster membership. Interestingly, we note that the clusters appear to correspond closely to the geographical distribution of the grid stations, suggesting that location affects the response of a station to a fault. Further investigation is required to uncover the deeper significance of these intra-class subgroups in effectively responding to faults, as well as any potential connection to fault severity.

\section{Conclusions and Future Work} \label{Conclusions and Future Work}
An early warning system prototype to detect, classify, and diagnose BPA electric grid faults was introduced. An efficient and effective anomaly detection method using quartiles was employed to detect the presence of a fault. Following this, the autocorrelation function was used to generate a compact, highly discriminative feature representation of a time series. Comparing the support vector machine, random forest, and artificial neural network classifier models, the random forest classifier was observed to be the superior technique in terms of accuracy, consistency, robustness to sparse training data, and excellent performance on incomplete time series for preemptive fault classification. Finally, intra-class clustering was conducted using partitioning around medoids with the distance metric defined by dynamic time warping . This uncovered five clusters that correlated closely with geographic location. 

Future work will focus on gaining deeper understanding of the implications of intra-class subgroups, including the possibility of fault severity estimation. Variables thus far unused, namely  voltage and phase angle can be used to provide additional information and granularity in the clustering process. In addition, developing more effective spatiotemporal visualizations of time series in the BPA grid is of crucial importance in enabling operators to assess and act quickly in potentially disastrous situations. These may include real-time heat maps of outlier frequency, rate of change, and other metrics. Additional data such as weather related parameters (temperature, humidity, etc.) could also be incorporated to aid the classification of faults with near-identical time series.

\section{Acknowledgements}
The authors would like to thank Joyce Luo, Jeremy Nguyen, and Zachary Yung for their contributions to early stages of this work. Thanks to Ms. Erica Bailey at Pittsford-Mendon High School for initiating this collaboration. Finally, thanks to Professor Esa Rantanen from the RIT department of Psychology for introducing us to the BPA data and the themes therein.


\begin{thebibliography}{10}

\bibitem{Farshad:2017:1}
Sadeh~J. Farshad, M.
\newblock Fault locating in high voltage transmission lines based on harmonic
  components of one-end voltage using random forests.
\newblock {\em Iranian Journal of Electrical \& Electronic Engineering}, 9,
  2013.

\bibitem{Hasan:2017:1}
Eboule P. Twala~B. Hasan, A.
\newblock The use of machine learning techniques to classify power transmission
  line fault types and locations.
\newblock {\em International Conference on Optimization of Electrical and
  Electronic Equipment}, 2017.

\bibitem{Malhotra:2012:1}
Jain~A. Malhotra, R.
\newblock Fault prediction using statistical and machine learning methods for
  improving software quality.
\newblock {\em Journal of Information Processing Systems}, 8, 2012.

\bibitem{Tayeb:2013:1}
M.~Tayeb.
\newblock Fault detection in power systems using artificial neural network.
\newblock {\em American Journal of Engineering Research}, 2, 2013.

\bibitem{FERREIRA:2018:1}
Eduardo~F. Ferreira and J.~Dionísio Barros.
\newblock Faults monitoring system in the electric power grid of medium
  voltage.
\newblock {\em Procedia Computer Science}, 130:696 -- 703, 2018.
\newblock The 9th International Conference on Ambient Systems, Networks and
  Technologies (ANT 2018) / The 8th International Conference on Sustainable
  Energy Information Technology (SEIT-2018) / Affiliated Workshops.

\bibitem{Dalstein:1995:1}
T.~Dalstein and B.~Kulicke.
\newblock Neural network approach to fault classification for high-speed
  protective relaying.
\newblock {\em IEEE Trans Power Deliv}, 4, 1995.

\bibitem{Jamil:2015:1}
Majid Jamil, Sanjeev~Kumar Sharma, and Rajveer Singh.
\newblock Fault detection and classification in electrical power transmission
  system using artificial neural network.
\newblock {\em SpringerPlus}, 4(1):334, Jul 2015.

\bibitem{Bhattacharya:2017:1}
Biswarup Bhattacharya and Abhishek Sinha.
\newblock Intelligent fault analysis in electrical power grids.
\newblock {\em CoRR}, abs/1711.03026, 2017.

\bibitem{doi:10.1177/1541931213601976}
Esa~M. Rantanen, Ernest Fokoué, Kathleen Gegner, Jacob Haut, and Thomas~J.
  Overbye.
\newblock Data properties underlying human monitoring performance.
\newblock {\em Proceedings of the Human Factors and Ergonomics Society Annual
  Meeting}, 61(1):1711--1715, 2017.

\bibitem{Vapnik:97:1}
V.~N. Vapnik, S.~E. Golowich, and A.~J. Smola.
\newblock {\it Support Vector method for function approximation, regression
  estimation and signal processing}.
\newblock In M.~I.~Jordan M.~C.~Mozer and T.~Petsche, editors, {\em Advances in
  Neural Information Processing Systems}, number~9. MIT Press, 1997.

\bibitem{Breiman:2001:1}
L.~Breiman.
\newblock Random forests.
\newblock {\em Machine Learning}, 45:5--32, 2001.

\bibitem{rfpic}
Will Koehrsen.
\newblock Random forest simple explanation, 2017.

\bibitem{BishopCM}
C~M Bishop.
\newblock {\em Neural Networks and Pattern Recognition}.
\newblock Oxford University Press, first edition, 1995.

\bibitem{Sakoe:79:1}
H.~Sakoe and S.~Chiba.
\newblock Dynamic programming algorithm optimization for spoken word
  recognition.
\newblock {\em IEEE Transactions on Acoustics, Speech, and Signal Processing},
  26:43--49, 1978.

\bibitem{kaufman:87:1}
L.~Kaufman and P.J. Rousseeuw.
\newblock Clustering by means of medoids.
\newblock 1987.

\bibitem{Lloyd:82:1}
S.~Lloyd.
\newblock Least squares quantization in pcm.
\newblock {\em IEEE Transactions on Information Theory}, 28:129--137, 1982.

\bibitem{Degerine:03:1}
S.~D\'egerine and S.~Lambert-Lacroix.
\newblock Characterization of the partial autocorrelation function of
  nonstationary time series.
\newblock {\em Journal of Multivariate Analysis}, 87:46--59, 2003.

\bibitem{Stoica:05:1}
P.~Stoica and R.~Moses.
\newblock {\em Spectral Analysis of Signals}.
\newblock Prentice Hall, 2005.

\end{thebibliography}

\end{document}